\documentclass[twocolumn, switch]{article}
\usepackage{preprint}
\usepackage{algorithm}
\usepackage{algorithmic}
\usepackage{amsmath, amsthm, amssymb, amsfonts}

\usepackage[numbers,square]{natbib}
\usepackage[utf8]{inputenc}
\usepackage[T1]{fontenc}
\usepackage{xcolor}
\usepackage[colorlinks = true,
            linkcolor = purple,
            urlcolor  = blue,
            citecolor = cyan,
            anchorcolor = black]{hyperref}
\usepackage{booktabs}
\usepackage{nicefrac}
\usepackage{microtype}
\usepackage{lineno}
\usepackage{float}
\usepackage{lipsum}
\usepackage{newfloat}
\DeclareFloatingEnvironment[name={Supplementary Figure}]{suppfigure}
\usepackage{sidecap}
\sidecaptionvpos{figure}{c}

\usepackage{titlesec}
\titlespacing\section{0pt}{12pt plus 3pt minus 3pt}{1pt plus 1pt minus 1pt}
\titlespacing\subsection{0pt}{10pt plus 3pt minus 3pt}{1pt plus 1pt minus 1pt}
\titlespacing\subsubsection{0pt}{8pt plus 3pt minus 3pt}{1pt plus 1pt minus 1pt}

\usepackage{eso-pic}
\usepackage{tikz}
\usepackage{xcolor}
\PassOptionsToPackage{colorlinks=true,linkcolor=gray,urlcolor=gray}{hyperref}

\usepackage{titling}
\usepackage{footmisc}
\usepackage{multirow}
\usepackage{subcaption}
\usepackage[table]{xcolor}
\usepackage{amsmath}

\newcommand{\Author}[2]{\textbf{#1}\textsuperscript{#2}}

\title{SSMamba: A Self-Supervised Hybrid State Space Model for Pathological Image Classification}

\author{
  \Author{Enhui Chai}{1} \and
  \Author{Sicheng Chen}{2}\and
  \Author{Tianyi Zhang}{3}\and
  \Author{Xingyu Li}{1} \and
  \Author{Tianxiang Cui}{4}
}

\date{
  \textsuperscript{1}School of Computer Science, Northwest University \\
  \textsuperscript{2}PuzzleLogic Pte Ltd, Singapore 229594, Singapore \\
  \textsuperscript{3}Department of Electrical \& Computer Engineering, National University of Singapore \\
  \textsuperscript{4}School of Computer Science, University of Nottingham Ningbo China \\
  [1em]
  \footnotesize \textbf{Corresponding author:} Tianxiang Cui\texttt{<tianxiang.cui@nottingham.edu.cn>} \\
}

\begin{document}
\twocolumn[\begin{@twocolumnfalse}
\maketitle
\thispagestyle{empty}
\begin{abstract}
\label{abstract}
Pathological diagnosis is highly reliant on image analysis, where Regions of Interest (ROIs) serve as the primary basis for diagnostic evidence, while whole-slide image (WSI)-level tasks primarily capture aggregated patterns. To extract these critical morphological features, ROI-level Foundation Models (FMs) based on Vision Transformers (ViTs) and large-scale self-supervised learning (SSL) have been widely adopted. However, three core limitations remain in their application to ROI analysis: (1) cross-magnification domain shift, as fixed-scale pretraining hinders adaptation to diverse clinical settings; (2) inadequate local-global relationship modeling, wherein the ViT backbone of FMs suffers from high computational overhead and imprecise local characterization; (3) insufficient fine-grained sensitivity, as traditional self-attention mechanisms tend to overlook subtle diagnostic cues. To address these challenges, we propose SSMamba, a hybrid SSL framework that enables effective fine-grained feature learning without relying on large external datasets. This framework incorporates three domain-adaptive components: Mamba Masked Image Modeling (MAMIM) for mitigating domain shift, a Directional Multi-scale (DMS) module for balanced local-global modeling, and a Local Perception Residual (LPR) module for enhanced fine-grained sensitivity. Employing a two-stage pipeline, SSL pretraining on target ROI datasets followed by supervised fine-tuning (SFT), SSMamba outperforms 11 state-of-the-art (SOTA) pathological FMs on 10 public ROI datasets and surpasses 8 SOTA methods on 6 public WSI datasets. These results validate the superiority of task-specific architectural designs for pathological image analysis.
\end{abstract}

\vspace{0.35cm}
\end{@twocolumnfalse}]
\section{Introduction}
\label{Introduction}
Pathological diagnosis is pivotal in clinical cancer management, as it relies on the interpretation of pathological images to assess disease progression and guide treatment planning~\cite{2}. Within these digital slides, Regions of Interest (ROIs) are specific sub-regions containing critical diagnostic indicators (e.g., abnormal cells, tissue structures) and serve as the primary focus for detailed examination, given their direct relevance to disease characteristics. Deep learning exhibits robust representation learning capabilities in this field~\cite{4,5}; however, its success is often contingent on large-scale annotated datasets, which are notoriously scarce and costly to acquire in pathology. To mitigate this dependency, Self-Supervised Learning (SSL)~\cite{ssl} has emerged as a powerful paradigm, enabling the extraction of general visual representations from unlabeled data via Contrastive Learning (CL)~\cite{cl} and Masked Image Modeling (MIM)~\cite{mim}. While these strategies have narrowed the performance gap with supervised learning across diverse downstream tasks~\cite{8}, they often fail to capture domain-specific distributions or bridge the semantic gap between upstream natural image datasets (e.g., ImageNet~\cite{6,7}) and downstream pathological images, yielding suboptimal representations.

To address this gap, large-scale SSL on diverse pathological cohorts has driven the development of pathological foundation models (FMs)~\cite{plism}. Predominantly based on Vision Transformer (ViT)-derived architectures~\cite{ViT,Swin}, these models have been widely adopted for their cross-task and cross-dataset generalizability. Nevertheless, recent studies have revealed a critical insight: even when trained on extensive datasets, general-purpose FMs often reach a performance plateau, whereas task-specific models with domain-aware designs consistently outperform them~\cite{sun2024pathasst, raghu2019transfusion}. These observations suggest that embedding pathology-aware inductive biases, coupled with targeted in-domain SSL and lightweight task-specific adaptation, may more effectively leverage limited labels and heterogeneous cohorts. Considering the unique morphological characteristics of pathological images, we identify three critical challenges:

\noindent \textbf{Magnification and image domain shift.} 
Variations in magnification, scanning hardware, and staining protocols induce significant distribution shifts~\cite{11,12}. Most mainstream FMs adopt Transformer architectures pretrained at a fixed magnification (e.g., UNI uses $0.5 mpp$, with $256$ pixels corresponding to $\approx 128 \mu m$ ROIs). This single-scale paradigm fails to accommodate the diverse magnification levels required in clinical practice, leading to domain mismatch when applied to high- or low-magnification ROIs. Consequently, downstream tasks cannot fully exploit the pathological knowledge embedded in FMs.

\noindent \textbf{Inadequate local-global integration.} 
Accurate diagnosis depends on both capturing subtle cellular morphology and integrating structural patterns across the entire visual field. CNN-based methods~\cite{lerousseau2020weakly} possess strong local inductive biases but struggle to model long-range structural relationships; in contrast, Transformer-based methods~\cite{stegmuller2023scorenet} can model long-range dependencies but incur prohibitive computational costs on high-resolution images and often fail to characterize continuous local transitions. State Space Models (SSMs)~\cite{ssm} offer a promising alternative by treating image tokens as continuous sequences with linear complexity. However, existing Mamba-based models~\cite{mamba} still suffer from autoregressive bias~\cite{mambaout}, which conflicts with the non-sequential and spatially entangled nature of pathological tissues.

\noindent \textbf{Fine-grained modeling misalignment.} 
Pathological analysis is inherently a fine-grained recognition task. Critical diagnostic clues often reside in subtle variations in glandular contours, nuclear pleomorphism, and texture boundaries. Traditional self-attention mechanisms prioritize global relationships and lack the inherent inductive bias required for local pattern modeling, resulting in insufficient sensitivity to these subtle yet decisive morphological cues~\cite{9}.

These findings indicate that the distinctive attributes of histopathological images—spatial heterogeneity and multiscale dependencies—demand pathology-aware inductive biases and training regimes that prioritize in-domain robustness over mere pretraining scale. Accordingly, we propose SSMamba, a two-stage framework for ROI classification. Our approach combines targeted in-domain SSL with task-specific fine-tuning, integrating three core components: (i) Mamba Masked Image Modeling (MAMIM) for robust visual initialization; (ii) the SSMamba encoder, which achieves joint local-global modeling with linear complexity; and (iii) Local Perception Residual (LPR) and Directional Multi-scale (DMS) modules to explicitly enhance fine-grained sensitivity. Our main contributions are:

\noindent \textbf{In-domain SSL fine-tuning integrated framework.} 
We propose an end-to-end framework that combines MAMIM initialization with SSMamba encoder fine-tuning for downstream ROI datasets. MAMIM adopts MAE-style masking and reconstruction tailored to SSM architectures, directing the model to critical structural signals (e.g., cell morphology, tissue topology) and mitigating cross-magnification domain shift. The hierarchical SSMamba encoder enables long-range context modeling via sequential scanning and hidden-state maintenance, preserving multi-scale structures. It achieves full integration of nuclear- or gland-level features with linear complexity, realizing synergy between initialization and fine-tuning.

\noindent \textbf{Enhanced fine-grained sensitivity.} 
We introduce the LPR and DMS modules to refine local patterns. LPR employs depthwise convolutions for spatial filtering to enhance boundary sensitivity, while DMS decomposes the Mamba backbone into multi-branch paths to capture local neighborhood relationships, effectively amplifying subtle morphological differences.

\noindent \textbf{State-of-the-art (SOTA) performance.} 
Across 10 ROI benchmarks, SSMamba outperforms 11 pathological FMs; on CAM16, it achieves +5.09\% Acc and +3.83\% F1 over Gigapath, validating the effectiveness of the proposed components. SSMambaMIL demonstrates superior performance across 6 WSI-based downstream tasks on 6 datasets, achieving the best results in 12 out of 17 metrics. Compared with other large-parameter algorithms, SSMamba only has 25.3M parameters.
\section{Related Work}
\label{Related Work}
\subsection{Self-Supervised Learning in Computational Pathology} SSL has become a cornerstone of representation learning in computational pathology, where expert annotations are notoriously scarce and costly. Contemporary SSL paradigms are primarily categorized into CL and MIM.

\noindent \textbf{Contrastive Learning.} 
Early advancements in pathological SSL were fueled by CL, which centers on instance discrimination. CTransPath~\cite{wang2022transformer} proposed a seminal transformer-based CL approach, utilizing context-aware contrastive objectives to extract discriminative features from unlabeled whole-slide images (WSIs) and achieving state-of-the-art performance across diverse downstream tasks.

\noindent \textbf{Masked Image Modeling.} 
Recently, MIM has gained prominence due to its reconstruction-centric objectives and inherent robustness to label noise. Modern pathology FMs, such as UNI~\cite{uni} and Virchow2~\cite{virchow2}, leverage the DINOv2 framework with teacher-student distillation and tissue-specific masking to preserve critical morphology. Furthermore, GigaPath~\cite{gigapath} scaled this paradigm to a billion-parameter model pretrained on 15 million patches, demonstrating that massive-scale MIM can significantly improve generalization across diverse cancer types. These models generally outperform earlier medical-specific MIM variants, such as SelfMedMAE~\cite{selfmedmae}, in modeling complex tissue architectures.

\noindent \textbf{Multimodal SSL.} 
To integrate domain-specific knowledge, multimodal SSL aligns visual features with clinical text data. MUSK~\cite{musk} leverages vision-language alignment to exploit paired histopathology reports, while CONCH~\cite{conch} employs language-supervised pretraining on specialized medical corpora to inject diagnostic semantics into image embeddings, enabling more interpretable feature extraction.

\subsection{State Space Models in Computer Vision.} The Mamba architecture~\cite{mamba}, rooted in structured state space models (SSMs), has emerged as a robust alternative to Transformers, attributed to its linear computational complexity and superior long-range dependency modeling. VMamba~\cite{vmamba} adapted this architecture for 2D vision via a Cross-Scan Module, which serializes images into directional sequences to balance spatial modeling capability with computational efficiency.

Despite these merits, standard Mamba-based models encounter unique challenges in pathological ROI analysis:

\noindent \textbf{Causal and autoregressive bias:} 
The sequential scanning mechanism of Mamba often conflicts with the spatially entangled and non-sequential nature of tissues~\cite{mambaout}, where diagnostic signals (e.g., cell clusters) do not adhere to a fixed 1D order. 

\noindent \textbf{Structural inflexibility:} 
Fixed scanning paths cannot dynamically adapt to the hierarchical and heterogeneous characteristics of tissues, frequently failing to prioritize clinically relevant structures.

\noindent \textbf{Sensitivity to artifacts:} 
Linear scanning may inadvertently amplify staining variations and translational noise, stemming from the absence of explicit pathology-aware inductive biases.

While recent adaptations (e.g., NaMA-Mamba~\cite{Nama}, SpineMamba~\cite{zhang2025spinemamba}) have been developed for 3D or endoscopic data, they fail to address the specific demands of fine-grained ROI classification—such as cross-magnification shift and subtle glandular morphological features. This motivates our development of SSMamba, which tailors the SSM paradigm to the unique nuances of histopathology.

\begin{figure*}[ht]
\centering
\includegraphics[width=\linewidth]{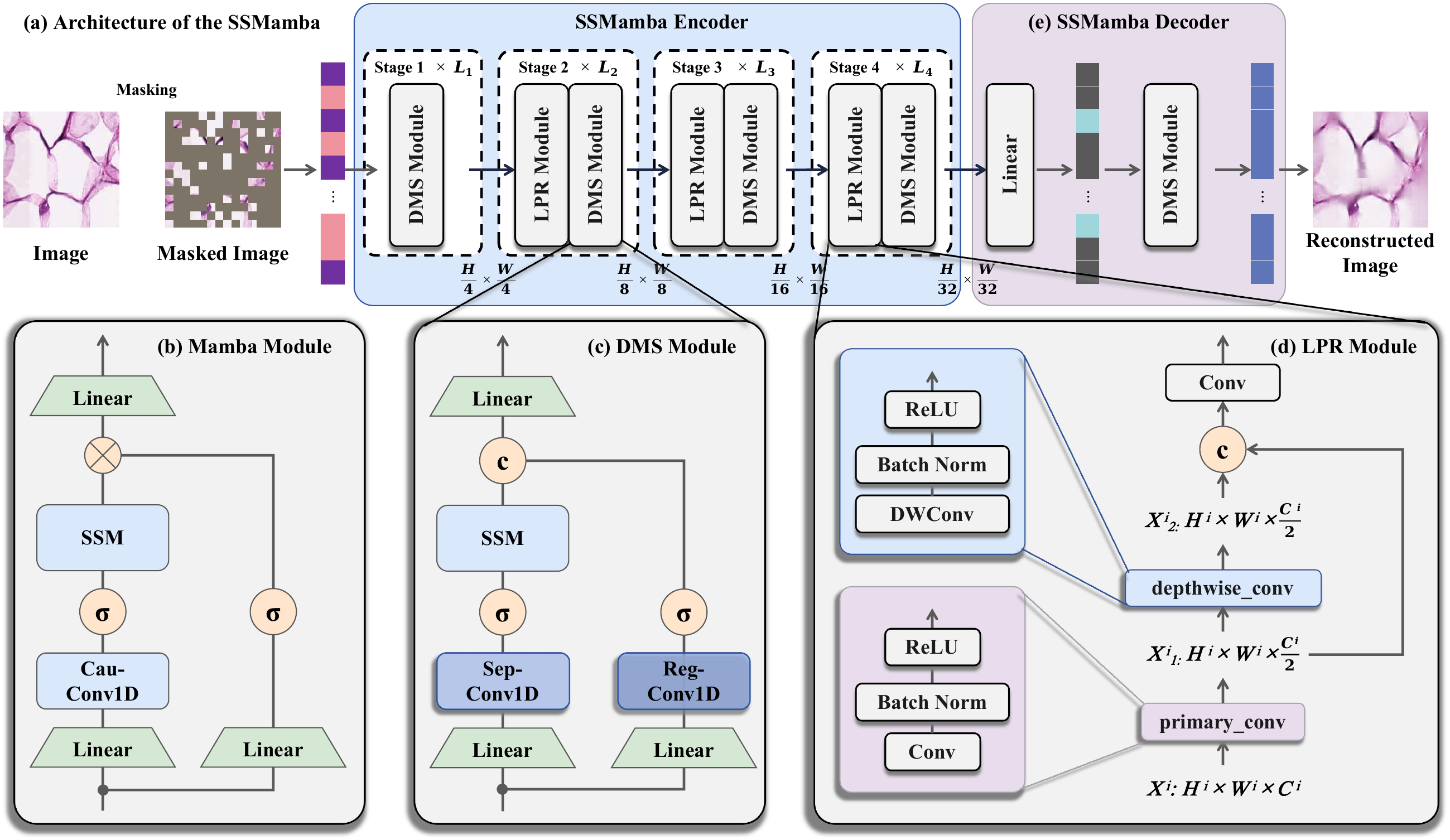}
\caption{The Architecture of the SSMamba. (a) Overall architecture of SSMamba. (b) The original Mamba Module. (c) The proposed DMS Module. (d) The proposed LPR Module. (e) The SSMamba decoder for MAMIM.}
\label{fig1}
\end{figure*}
\section{Methods}
\label{Methods}
\subsection{Overall Architecture}
Existing pathological FMs are predominantly based on ViT architectures and generic computer vision SSL paradigms (e.g., MAE, CL). However, their designs often overlook pathology-specific characteristics, leading to suboptimal feature representations for pathological image analysis. To bridge this gap, we propose SSMamba, a task-specific two-stage framework (Fig.~\ref{fig1}(a)) featuring a four-stage hierarchical encoder ($L_1$–$L_4$) customized for pathological scenarios. This architecture is designed to tackle three core challenges inherent to ROI analysis.

To mitigate cross-magnification domain shift, we propose the MAMIM module (Sec.~\ref{MAMIMM}), which reduces domain discrepancies and suppresses artifact-prone regions. For an input image $X \in \mathbb{R}^{H \times W \times 3}$, a 75\% random masking strategy ($m_r = 0.75$) is adopted, where $m_r \cdot (H \cdot W)$ patches are replaced with learnable mask tokens $X_m$. This design preserves contextual cues through unmasked anchor patches, ensuring compatibility with downstream diagnostic tasks. For enhanced local-global relationship modeling, a hierarchical encoder (SSMamba encoder) based on the SSM is developed. This encoder serializes ROI features for long-range context modeling, while preserving multi-scale structures through layer-wise fusion of nuclear- and gland-level features. Specifically, SSMamba incorporates a DMS module (Sec.~\ref{DMSModule}) to enable full token interaction. Diverging from conventional MIM designs, the DMS adopts a pyramidal architecture where each stage $L_k$ refines and aggregates multi-scale representations. Moreover, the DMS module facilitates concurrent local-global information flow via bidirectional state propagation, enabling effective feature extraction ranging from fine-grained cellular patterns ($L_1$) to high-level tissue topology ($L_4$).

Conventional coordinate-based encodings (e.g., ViT’s absolute positional embeddings) are vulnerable to tissue shifts and rotations. To resolve this, we introduce the LPR module (Sec.~\ref{LPRModule}), which replaces these encodings with dynamic depthwise convolutions to enable spatially invariant modeling of architectural patterns. Formally, the hierarchical encoding process is defined as:
\begin{equation}
\mathbf{F}_{k+1} = \text{DMS}_k(\text{LPR}_k(\mathbf{F}_k)), \quad k \in {1,2,3,4}
\end{equation}
where $\mathbf{F}$ denotes features. Each stage $k$ progressively downscales the spatial resolution while increasing channel depth, thus preserving high diagnostic fidelity.

For ROI classification, the model's head consists of a global average pooling layer applied to the final feature map, followed by a single fully-connected (linear) layer that projects the features to the number of target classes. For WSI classification tasks, the model's prediction head adopts a multi-task learning (MTL) design where outputs from SSMambaMIL are fed into multiple task-specific heads, with a modular pipeline automatically constructing corresponding classification or regression layers based on task configurations, supporting flexible gradient computation for tasks with available indicators and joint learning via aggregated task losses (cross-entropy loss for classification and mean absolute error loss for regression).

Overall, SSMamba requires no auxiliary annotations or preprocessing, exhibiting strong adaptability across diverse pathological datasets. It prioritizes domain-aware robustness over generic large-scale pretraining.

\subsection{MAMIM-based Encoder and Decoder}
\label{MAMIMM}
Cross-magnification domain shift impairs model generalization and elevates the risk of misdiagnosis in pathological image analysis. To address this, we propose MAMIM, whose core innovation lies in replacing all vanilla ViT blocks in the standard MAE encoder-decoder framework with customized SSMamba blocks (integrating DMS and LPR modules). This targeted replacement resolves the mismatch between generic MAE and pathological image characteristics, enabling effective mitigation of cross-magnification domain shift.

MAMIM follows MAE’s classic masking-reconstruction paradigm but incorporates architecture-specific optimizations for pathology: For an input image $X \in \mathbb{R}^{H \times W \times 3}$, it is first partitioned into non-overlapping patches, a 75\% masking ratio is applied, and the masked sequence is fed into the SSMamba-based encoder (stacked SSMamba blocks) for multi-scale feature extraction, which distinct from MAE’s ViT-based encoder. The integrated modules of the SSMamba block are key to distinguishing it from MAE’s ViT blocks: (1) Instead of ViT’s absolute positional encoding (susceptible to staining artifacts and tissue shifts), the LPR module employs dynamic depthwise convolutions to achieve translation invariance, adapting to pathological image variations. (2) Unlike ViT’s unidirectional self-attention (inadequate for capturing tissue interactions), the DMS module enables bidirectional context aggregation and fuses parallel convolutional features, effectively capturing global tissue topology (e.g., tumor-stroma boundaries) and local cellular patterns. These designs ensure the encoder extracts domain-invariant, diagnostically relevant features, avoiding MAE’s tendency to learn magnification-specific artifacts.

Fig.~\ref{fig1}(e) illustrates the architecture of the SSMamba decoder, a core component of the MAMIM pretraining framework. For decoding, MAMIM adopts an SSMamba-based decoder (instead of MAE’s ViT-based decoder) to maintain architectural consistency with the encoder. Leveraging the DMS module’s direction-aware propagation, the decoder recovers masked patches using contextual cues from unmasked anchors and outputs a reconstructed image. Compared to MAE’s generic decoder, this SSMamba-based design enhances the capture of complex pathological structures, ensuring consistent feature learning during pretraining. Facilitated by the encoder-decoder’s architectural advantages over MAE, the reconstruction task drives the model to learn consistent pathological tissue patterns across magnifications, fundamentally alleviating domain shift. These architectural improvements endow MAMIM with superior pathological adaptability and efficient cross-magnification domain alignment, advantages lacking in standard MAE, which relies on generic ViT blocks, absolute positional encoding, and unidirectional self-attention that are mismatched with pathological image characteristics.

\begin{table}[htbp]
\centering
\caption{Architectural Comparison: Mamba Module in VMamba vs. DMS Module in SSMamba}
\label{tab:mamba_comparison}
\resizebox{\linewidth}{!}{
\begin{tabular}{lll}
\toprule
\textbf{Feature} & \textbf{Mamba} & \textbf{DMS} \\
\midrule
Token Mixing & Unidirectional & Bidirectional \\
MAE Compatibility & Limited & Full \\
Spatial Processing & SSM & SSM + Parallel Conv \\
Pathology Optimization & Foundation model & Tissue integrity \\
Activation & GeLU & SiLU \\
\bottomrule
\end{tabular}}
\end{table}

\subsection{DMS Module}
\label{DMSModule}
To address inherent limitations of the original Mamba module (Fig.~\ref{fig1}(c)) in pathological image analysis tasks, we redesign it into the DMS module. The DMS module targets three core flaws of vanilla Mamba, with corresponding optimized designs tailored for pathological image analysis.

First, DMS overcomes the unidirectional constraint of vanilla Mamba, which relies on causal convolution (Cau-Conv1D) to enforce left-to-right sequence modeling (Eq.~\ref{eq:causal_conv}):
\begin{equation}
X_{\text{causal}}[t] = \sum_{i=0}^{k} W[i] \cdot X[t-i]
\label{eq:causal_conv}
\end{equation}
where $X_{\text{causal}}[t]$ is the output at position $t$, and $k$ denotes the kernel size, $W[i]$ is the learnable weight, and $X[t-i]$ is the input feature at position $t-i$. This unidirectional design fails to capture bidirectional spatial context, which is critical for pathology, as malignant patterns (e.g., tumor-stroma boundaries) depend on mutual interactions between adjacent regions.

To resolve this, we replace Cau-Conv1D with bidirectional depthwise separable convolution (Sep-Conv1D), consisting of per-channel depthwise filtering and cross-channel pointwise mixing (Eq.~\ref{eq:sep_conv}):
\begin{equation}
\begin{aligned}
&\text{Depthwise:} \quad X_{\text{dw}}[c,t] = \sum_{i=-k/2}^{k/2} W_{\text{dw}}[c,i] \cdot X[c,t+i] \\
&\text{Pointwise:} \quad X_1' = W_{\text{pw}} \cdot X_{\text{dw}}
\end{aligned}
\label{eq:sep_conv}
\end{equation}
where $X_{\text{dw}}[c,t]$ is the depthwise output at channel $c$ and position $t$, $W_{\text{dw}}[c,i]$ is the channel-specific depthwise weight, and $W_{\text{pw}}$ denotes the pointwise weight matrix. This design enables centered bidirectional context aggregation to capture global tissue topology (e.g., tumor-stroma interfaces) while reducing the parameter count by a factor of $1/k + 1/C$ compared to standard convolution, ensuring efficiency without compromising expressiveness.

Second, DMS remedies the lack of parallel spatial dependencies in vanilla Mamba. The sequential nature of SSM neglects parallel spatial interactions, which are essential for capturing local cellular patterns. We thus introduce a symmetric convolutional branch with regular 1D convolution and SiLU activation (Eq.~\ref{eq:conv_branch}):
\begin{equation}
X_2 = \sigma\left(\text{Reg-Conv1D}\left(\text{Linear}_{C \to C/2}(X_{in})\right)\right)
\label{eq:conv_branch}
\end{equation}
where $\sigma(\cdot)$ is the SiLU activation, $\text{Linear}_{C \to C/2}$ projects channels from $C$ to $C/2$, and $X_{in}$ is the input feature map. This branch processes tokens concurrently, enhancing local discriminability and robustness to visual degradation—critical for distinguishing visually similar cancer subtypes.

Third, DMS resolves the incompatibility between Mamba’s autoregressive bias and the non-autoregressive MAE framework, which induces training instability. We fuse the bidirectional SSM pathway and convolutional pathway via channel-split concatenation to maintain efficiency (Eq.~\ref{eq:fusion}):
\begin{equation}
\begin{aligned}
&X_1 = \text{Scan}\left(\sigma\left(\text{Sep-Conv1D}\left(\text{Linear}_{C \to C/2}(X_{in})\right)\right)\right) \\
&X_{out} = \text{Linear}_{C \to C}\left(\left[X_1 \oplus X_2\right]\right)
\end{aligned}
\label{eq:fusion}
\end{equation}
where $\text{Scan}$ denotes Mamba’s selective scan operation (formulated as $h_t = \bar{A}_t h_{t-1} + \bar{B}_t x_t, \, y_t = C_t h_t$, where $\bar{A}_t = \exp(-\Delta_t A)$ and $\bar{B}_t = \Delta_t B$ are discretized via zero-order hold), $\oplus$ denotes channel-wise concatenation, and $\text{Linear}_{C \to C}$ restores the channel dimension. The $C \to C/2$ channel-split design ensures full token interaction while preserving the parameter count of the original Mamba block.

In summary, the DMS module extends Mamba’s sequence modeling capability to pathological spatial feature extraction by integrating bidirectional recurrence, parallel convolution, and MAE-compatible fusion. Table~\ref{tab:mamba_comparison} summarizes its architectural advantages over the Mamba module in VMamba. The pseudocode of DMS is provided in Algorithm~\ref{alg1}.

\begin{algorithm}[ht] 
\caption{DMS Module}
\label{alg1}
\begin{algorithmic}[1]
\REQUIRE Input sequence $X_{in} \in \mathbb{R}^{L \times d}$, sequence length $L$, hidden dimension $d$
\ENSURE Output sequence $X_{out} \in \mathbb{R}^{L \times d}$

\STATE \textbf{Parameters:}
\STATE \quad $A \in \mathbb{R}^{N \times N}$ \COMMENT{State transition matrix}
\STATE \quad $B, C \in \mathbb{R}^{d \times N}$ \COMMENT{Projection matrices}
\STATE \quad $K \in \mathbb{R}^{k}$ \COMMENT{Depthwise convolution kernel}
\STATE \quad $\gamma, \beta \in \mathbb{R}^{d}$ \COMMENT{Layer normalization parameters}

\STATE \textbf{Forward Pass:}

\STATE \textbf{Step 1: Input Projection}
\STATE \quad $\Delta \leftarrow \text{Softplus}(\text{Linear}_{\Delta}(X_{in}))$
\STATE \quad Split $X_{\text{in}}$ into two branches: $X_{\text{in1}}, X_{\text{in2}}$
\STATE \quad $B_{in} \leftarrow \text{Linear}_B(X_{\text{in1}})$ 
\STATE \quad $C_{in} \leftarrow \text{Linear}_C(X_{\text{in2}})$ 
\STATE \quad $A_{\text{scaled}} \leftarrow \text{exp}(-\Delta \cdot A)$ 
\STATE \quad $X_{\text{proj1}} \leftarrow \text{Linear}_{C \to C/2}(X_{in1})$
\STATE \quad $X_{\text{proj2}} \leftarrow \text{Linear}_{C \to C/2}(X_{in2})$

\STATE \textbf{Step 2: Convolution}
\STATE \quad $X_{\text{conv1}} \leftarrow \text{Sep-Conv1D}(X_{\text{proj1}}, K)$ 
\STATE \quad $X_{\text{conv2}} \leftarrow \text{Reg-Conv1D}(X_{\text{proj2}})$
\STATE \quad $X_{\text{act1}} \leftarrow \text{SiLU}(X_{\text{conv1}})$
\STATE \quad $X_{\text{act2}} \leftarrow \text{SiLU}(X_{\text{conv2}})$

\STATE \textbf{Step 3: Selective Scan}
\STATE \quad $X_{\text{ssm}} \leftarrow \text{SSM}(A_{\text{scaled}}, B_{in}, \Delta)(X_{\text{act1}})$ 

\STATE \textbf{Step 4: Gating Mechanism}
\STATE \quad $X_{\text{gated}} \leftarrow  X_{\text{act2}} \oplus X_{\text{ssm}}$ 

\STATE \textbf{Step 5: Output Projection}
\STATE \quad $X_{\text{out}} \leftarrow \text{Linear}_{\text{out}}(X_{\text{gated}})$
\RETURN $X_{\text{out}}$
\end{algorithmic}
\end{algorithm}

\begin{table*}[htbp]
\centering
\caption{Embedding Method Comparison for Pathology Image Classification.}
\label{tab:pe_comparison}
\resizebox{\textwidth}{!}{
\begin{tabular}{lllll}
\toprule
\textbf{Characteristic} & \textbf{Linear PE (ViT)} & \textbf{Patch-Merge (Swin)} & \textbf{LPU (CMT)} & \textbf{LPR (Ours)}\\
\midrule
\textbf{Translation Invariance} & $\times$ & \checkmark & \checkmark & \checkmark \\
\textbf{Resolution Preservation} & \checkmark & $\times$ & \checkmark & \checkmark \\
\textbf{Local Feature Extraction} & $\times$ & $\times$ & \checkmark & \checkmark \\
\textbf{Stain Artifact Robustness} & $\times$ & $\times$ & $\times$ & \checkmark \\
\textbf{Computational Cost} & $O(N)$ & $O(N/4)$ & $O(k^2NC)$ & $O(k^2NC/2)$ \\
\textbf{Implementation} & Linear Projection & Patch Concatenation & Conv+ReLU & DWConv+Residual \\
\textbf{Gradient Propagation} & Standard & Limited & Residual & Multi-scale residual \\
\bottomrule
\end{tabular}}
\end{table*}

\subsection{LPR Module}\label{LPRModule}
Pathological image analysis presents unique spatial challenges: diagnostic features (e.g., nuclear morphology) demand translation invariance, while Hematoxylin and Eosin (H\&E) staining-induced artifactual patterns (e.g., dye diffusion, tissue folds) readily corrupt inappropriate positional encodings and amplify noise. Conventional positional embedding methods fail to meet these domain-specific requirements: ViT’s linear positional encoding introduces fixed coordinate biases, violating translation invariance; Swin’s Patch-Merge achieves hierarchical downsampling but aggressively compresses spatial information, exacerbating staining artifacts; the Local Perception Unit (LPU) in CMT~\cite{cmt} employs depthwise separable convolutions for local feature extraction but lacks explicit adaptation to pathological artifacts.

To tackle these issues, we propose the LPR module, a domain-adaptive positional encoding module tailored for pathological scenarios. The LPR module consists of three core sequential operations, with stage input $X^i \in \mathbb{R}^{H^i \times W^i \times C^i}$ (where $H^i, W^i$ are spatial dimensions and $C^i$ is channel count): First, pointwise convolution compresses channels while preserving fine-grained cellular details:
\begin{equation}
X_L = \text{ReLU}\left(\text{BN}\left(\text{Conv}_{C^i \rightarrow C^i/2}(X^i)\right)\right)
\end{equation}
where $\text{Conv}_{C^i \rightarrow C^i/2}$ denotes a pointwise convolution mapping channels from $C^i$ to $C^i/2$, $\text{BN}$ is batch normalization for training stability, and $\text{ReLU}$ is the activation function. This operation reduces computational overhead while retaining critical cellular features, yielding $X_L \in \mathbb{R}^{H^i \times W^i \times C^i/2}$.

Second, depthwise convolution (DWConv) is applied to $X_L$ to introduce translation-invariant local perception:
\begin{equation}
X_{DW} = \text{BN}\left(\text{DWConv}_{k \times k}(X_L)\right)
\end{equation}
where $k \times k$ kernel (typically $3 \times 3$) is applied independently to each channel, enabling localized feature extraction with spatially shared weights to enhance translation invariance. Compared to standard convolution, DWConv reduces the parameter cost from $O(k^2 \cdot (C^i/2)^2)$ to $O(k^2 \cdot C^i/2)$ by avoiding inter-channel mixing.

Third, residual fusion restores the original representation to stabilize training and enhance context flow:
\begin{equation}
X_{Final} = \text{Conv}_{C^i/2 \rightarrow C^i}\left(\text{ReLU}(X_{DW})\right) + X^i
\end{equation}
This residual pathway facilitates gradient propagation, preserves original spatial semantics, and fuses multi-scale local perception with global tissue context—effectively decoupling staining noise from diagnostic signals.

As summarized in Table~\ref{tab:pe_comparison}, LPR fundamentally improves pathology-oriented feature embedding by integrating DWConv-based implicit positional encoding with residual pathways. The pseudocode of the LPR module is provided in Algorithm~\ref{alg2}.

\begin{algorithm}[htbp]
\caption{LPR Module}
\label{alg2}
\begin{algorithmic}[1]
\REQUIRE $X^i, C_{\text{int}}, C_{\text{out}}, s, k, st$
\ENSURE $X_{Final}$
    \STATE $C_{\text{int}} \gets C_{\text{out}} / s$ \COMMENT{Intrinsic channels}
    \STATE $C_{\text{ghost}} \gets C_{\text{int}} \times (s - 1)$ \COMMENT{Ghost channels}
    \STATE \textbf{Step 1: Primary convolution}
    \STATE \quad  $X_L \gets \text{Conv}_{k\times k}(X^i)$ \COMMENT{Generate $C_{\text{int}}$ features}
    \STATE \quad  $X_L \gets \text{BN}(X_L)$
    \STATE \quad  $X_L \gets \text{ReLU}(X_L)$
    
    \STATE \textbf{Step 2: Depthwise convolution}
    \STATE \quad  $X_{DW} \gets \text{DWConv}_{3\times 3}(X_L)$ \COMMENT{Generate $C_{\text{ghost}}$ features}
    \STATE \quad  $X_{DW} \gets \text{BN}(X_{DW})$
    \STATE \quad  $X_{DW} \gets \text{ReLU}(X_{DW})$
    
    \STATE \textbf{Step 3: Concatenate}
    \STATE \quad  $X_{Final} \gets \text{Concat}(X_L, X_{DW}, \text{dim}=1)$
    \RETURN $X_{Final}$
\end{algorithmic}
\end{algorithm}
\section{Experiments}
\label{Experiments}
\subsection{Datasets}
\begin{table*}[htbp]
\centering
\caption{Dataset Details.}
\label{dataset}
\resizebox{\textwidth}{!}{
\begin{tabular}{llllll}
\toprule
\textbf{Dataset} & \textbf{Classes} & \textbf{Resolution (pixels)} & \textbf{Sample Number} & \textbf{Organ/Tissue} & \textbf{Scale} \\
\midrule
LaC          & 5 & $768\times768$   & 25000  & Colorectal  & Tissue \\
NCT          & 9 & $224\times224$   & 100000 & Colorectal  & Tissue \\
PBC          & 8 & $360\times363$   & 38938  & Blood       & Cellular \\
pRCC         & 2 & $2000\times2000$ & 1419   & Kidney      & Glandular \\
PAIP2019     & 2 & $384\times384$   & 2165   & Liver       & Tissue \\
CAM16        & 2 & $8000\times8000$ & 1081   & Lymph Nodes & Tissue \\
SIPa     & 5 & $384\times384$   & 1004   & Cervical    & Cellular \\
MHIST        & 2 & $224\times224$ & 3152 & Colorectal & Tissue \\
TCGA-CRC-MSI & 2 & $512\times512$ & 51916 & Colorectal & Tissue \\
Osteosarcoma & 3 & $1024\times1024$ & 1144 & Bone & Tissue \\
\bottomrule
\end{tabular}}
\end{table*}
To ensure the generalizability of our proposed framework, we selected 10 publicly available pathology ROI datasets covering diverse tissue types, pathological fields, and scale ranges: Lung and Colon Cancer (LaC)~\cite{lac}, NCT-CRC-HE-100K (NCT)~\cite{nct}, Peripheral Blood Cell (PBC)~\cite{pbc}, Papillary Renal Cell Carcinoma (pRCC)~\cite{prcc}, PAIP2019~\cite{paip2019}, CAMELYON16 (CAM16)~\cite{cam16}, SIPaKMeD (SIPa)~\cite{sipakmed}, MHIST~\cite{MHIST}, TCGA-CRC-MSI (CRC)~\cite{tcgadata} and Osteosarcoma (Ost)~\cite{osteosarcoma}. Details of these datasets are summarized in Table~\ref{dataset}. Meanwhile, we evaluated our proposed algorithm on 6 public WSI datasets: 1)For prostate cancer subtyping using the PANDA~\cite{panda} dataset, and 2) TCGA~\cite{tcgadata} datasets, sourced from The Cancer Genome Atlas (TCGA) project, provide publicly accessible, high-quality WSIs along with matched clinical and molecular data. These datasets span a broad spectrum of cancers and have become standard benchmarks for deep learning research in histopathology. Specifically, the following subsets are used: TCGA-ESCA, TCGA-BRCA, TCGA-CESC, TCGA-LGG, and TCGA-BLCA. The details of the dataset are in Dataset Details of Appendix.

\subsection{Implementation Details}
All experiments were conducted on a hardware setup consisting of 3 RTX A6000 GPUs and 1 RTX A5000 GPU, with Python 3.10.16, PyTorch 2.4.0, and CUDA 12.1.

For ROI classification tasks, all datasets were partitioned into training, validation, and test sets following a 7:1:2 split ratio. For fair comparison, all images were resized to $224 \times 224 \times 3$ prior to model training, followed by normalization to a standard distribution across each dataset. Consistent with the MAE framework, a 75\% masking ratio was adopted for pretraining, with a total of 100 pretraining epochs conducted across all datasets. The pretraining process employed the AdamW optimizer with a base learning rate of $5 \times 10^{-5}$, weight decay of 0.05, cosine learning rate decay scheduling, a batch size of 64, and a 10-epoch warmup phase; data augmentation was primarily based on Random Resized Crop. For fine-tuning, the AdamW optimizer and weight decay of 0.05 were retained, while a higher base learning rate ($1 \times 10^{-3}$), cosine learning rate decay, batch size of 8, 10-epoch warmup, and Mixup augmentation were adopted. Model performance was evaluated using three standard metrics: F1-score (F1), Accuracy (Acc), and Area Under the Curve (AUC).

For WSI classification tasks, models were trained for 100 epochs (20 epochs for warmup and 80 epochs for main training), employing the AdamW optimizer with an initial learning rate of $1 \times 10^{-4}$ and a cosine decay schedule annealing to $1 \times 10^{-6}$. Model performance was evaluated using the accuracy metric. Each WSI was divided into $224 \times 224$ tiles, with features represented by Gigapath embeddings. To mitigate sampling variance—especially for small datasets—15 independent test inferences with different random tile samplings were performed for each scenario, and the predictions were aggregated. A batch size of 4 was maintained across all WSI experiments. SOTA models and the training framework were implemented based on the UnPuzzle Benchmark~\cite{UnPuzzle}.

\subsection{Comparison with SOTA Methods}
To validate the ROI pathological image classification performance of SSMamba, comparative experiments were conducted against 11 SOTA methods on the 10 aforementioned ROI datasets. The comparative baselines include ViT~\cite{ding2023enhanced}, Swin-Transformer (Swin)~\cite{cai2023mist}, MAE~\cite{mae}, VMamba~\cite{zhang20252dmamba}, UNI~\cite{uni}, UNI2~\cite{uni}, MUSK~\cite{musk}, CONCH~\cite{conch}, CTransPath~\cite{wang2022transformer}, Prov-GigaPath (Gigapath)~\cite{gigapath}, and Virchow2~\cite{virchow2}.

For validating SSMamba's performance on WSI pathological image classification, comparative experiments were performed against 8 SOTA methods across 6 core downstream tasks on 6 WSI datasets. The comparative baselines consist of 2 baseline methods (SlideAve, SlideMax) and 6 SOTA methods (ABMIL~\cite{abmil}, CLAM~\cite{clam}, DSMIL~\cite{dsmil}, Gigapath~\cite{gigapath}, S4MIL~\cite{s4mil}, MambaMIL~\cite{mambamil}). SSMamba was fine-tuned on 6 datasets: PANDA, TCGA-ESCA, TCGA-BRCA, TCGA-CESC, TCGA-LGG, and TCGA-BLCA. Performance comparisons between SSMamba and the competing methods across these downstream tasks—covering diverse scenarios of WSI-based pathological diagnosis—are presented in Tab.~\ref{tab:WSI}.

\noindent \textbf{Classification tasks:} 
(1) I-Grade: Assessing prostate cancer aggressiveness on the PANDA dataset, where the ISUP Grade serves as a standardized classification of Gleason patterns to guide prognosis and treatment planning; (2) Grade: Evaluating tumor grade via cellular differentiation and structural pattern analysis on the TCGA-ESCA dataset; (3) IHC-HER2: Predicting HER2 expression status from WSIs of the TCGA-BRCA dataset to determine suitability for targeted therapy; (4) LymInv: Identifying cancer invasion into lymphatic or vascular channels on the TCGA-CESC dataset; (5) T-Stage: Predicting AJCC-defined tumor staging (reflecting local tumor progression) on the TCGA-LGG dataset.

\noindent \textbf{Regression task:}
OS Months: Estimating overall survival time from initial diagnosis on the TCGA-BLCA dataset, using Correlation (Corr) as the evaluation metric.

\subsubsection{ROI Pathological Image Classification}

\begin{table*}[htbp]
\centering
\caption{Performance Comparison of SSMamba with 11 SOTA Methods on 10 ROI Pathology Datasets. \textbf{Bold}: the best result. \underline{Underline}: the second best result.}
\begin{subtable}[ht]{\textwidth}
\centering
\caption{F1 Score Results Comparison of SSMamba with 11 SOTA Methods on 10 Pathology Datasets.}
\label{comparisonF1}
\resizebox{\textwidth}{!}{
\begin{tabular}{lcccccccccccc} 
\toprule
\textbf{Method} & \textbf{LaC} & \textbf{NCT} & \textbf{PBC} & \textbf{pRCC} & \textbf{PAIP2019} & \textbf{CAM16} & \textbf{SIPa} & \textbf{MHIST} & \textbf{CRC} & \textbf{Ost} & \textbf{Ave-F1} & \textbf{p-value} \\
\midrule
ViT & 89.92 & 96.39 & 95.14 & 90.04 & 89.88 & 87.87 & 93.79 & 77.25 & 77.55 & 88.46 & 88.63 & 0.000002 \\
Swin & 92.03 & 97.33 & 97.59 & 92.69 & 87.67 & \underline{90.73} & 94.77 & 77.45 & 79.25 & 89.64 & 89.92 & 0.000005 \\
MAE & 95.88 & 98.57 & \underline{98.47} & 87.17 & 76.24 & 89.80 & 90.33 & 76.79 & 77.80 & 89.52 & 88.06 & 0.000010 \\
VMamba & 90.40 & 90.80 & 87.19 & 84.20 & 79.39 & 84.40 & 95.22 & 76.87 & 82.84 & 90.71 & 86.20 & 0.000001 \\
UNI & \textbf{99.98} & \underline{99.51} & 98.07 & 87.50 & 95.61 & 85.86 & 78.00 & 81.81 & 88.74 & 94.35 & 90.94 & 0.000028 \\
UNI2 & 98.07 & 99.14 & 98.15 & \underline{97.13} & 91.54 & 85.87 & 77.96 & 82.56 & \underline{89.38} & \underline{96.61} & 91.64 & 0.000003 \\
MUSK & 98.72 & 98.46 & 97.39 & 89.85 & 92.79 & 82.65 & 78.00 & 82.06 & 85.22 & 94.35 & 89.95 & 0.000004 \\
CONCH & 99.70 & 99.29 & 97.44 & 90.17 & 80.54 & 83.33 & 90.50 & \underline{83.63} & 86.60 & 94.58 & 90.58 & 0.000008 \\
GigaPath & \textbf{99.98} & \textbf{99.73} & 98.21 & 92.89 & 94.46 & 88.04 & \underline{98.50} & \textbf{83.85} & 89.09 & 95.95 & 94.07 & 0.000127 \\
Virchow2 & \underline{99.90} & 99.36 & 98.26 & 94.68 & \underline{96.49} & 89.95 & 98.00 & 83.42 & 88.98 & 95.24 & \underline{94.43} & 0.000015 \\
CTransPath & 91.29 & 92.16 & 95.87 & 92.80 & 88.23 & 90.04 & 98.30 & 79.73 & 84.87 & 91.42 & 90.47 & 0.000001 \\
\rowcolor{gray!40} SSMamba & 98.47 & 99.38 & \textbf{99.54} & \textbf{97.87} & \textbf{98.17} & \textbf{91.87} & \textbf{100.00} & 83.19 & \textbf{89.80} & \textbf{97.31} & \textbf{95.56} & - \\
\bottomrule
\end{tabular}}
\end{subtable}

\vspace{1em}

\begin{subtable}[ht]{\textwidth}
\centering
\caption{Accuracy Results Comparison of SSMamba with 11 SOTA Methods on 10 Pathology Datasets.}
\label{comparisonAcc}
\resizebox{\textwidth}{!}{
\begin{tabular}{lcccccccccccc} 
\toprule
\textbf{Method} & \textbf{LaC} & \textbf{NCT} & \textbf{PBC} & \textbf{pRCC} & \textbf{PAIP2019} & \textbf{CAM16} & \textbf{SIPa} & \textbf{MHIST} & \textbf{CRC} & \textbf{Ost} & \textbf{Ave-Acc} & \textbf{p-value}\\
\midrule
ViT & 92.11 & 97.63 & 96.84 & 92.55 & 91.89 & 90.31 & 95.13 & 83.19 & 79.36 & 92.50 & 91.15 & 0.000001 \\
Swin & 93.61 & 97.57 & 96.93 & 92.03 & 92.10 & \underline{92.50} & 95.35 & 83.19 & 79.59 & 95.59 & 91.85 & 0.000002 \\
MAE & 98.49 & 98.71 & 97.33 & 88.98 & 78.20 & 92.24 & 93.27 & 84.78 & 79.04 & 93.81 & 90.49 & 0.000001 \\
VMamba & 92.13 & 91.57 & 85.33 & 86.39 & 81.67 & 86.69 & 96.80 & 85.33 & 79.07 & 90.71 & 87.57 & 0.000000 \\
UNI & \textbf{99.99} & 99.89 & 99.52 & 90.81 & 95.60 & 87.04 & 91.20 & 86.41 & 81.54 & 96.23 & 92.72 & 0.000005 \\
UNI2 & 98.66 & 99.24 & 99.51 & 90.88 & 95.65 & 87.87 & 91.60 & 87.40 & \underline{82.62} & \textbf{97.74} & 93.12 & 0.000003 \\
MUSK & 99.49 & 99.66 & 99.35 & 92.58 & 93.05 & 84.26 & 91.20 & 86.87 & 76.10 & 96.23 & 91.88 & 0.000001 \\
CONCH & 99.88 & 99.84 & 99.36 & 92.93 & 76.85 & 85.15 & 96.20 & \underline{88.10} & 77.82 & 96.39 & 91.25 & 0.000001 \\
GigaPath & \textbf{99.99} & \textbf{99.94} & \underline{99.55} & 94.70 & 94.21 & 88.42 & \underline{99.40} & \textbf{88.22} & 81.99 & 97.30 & 94.37 & 0.000018 \\
Virchow2 & \underline{99.96} & \underline{99.86} & \textbf{99.56} & \underline{96.11} & \underline{96.53} & 90.28 & 99.20 & 87.57 & 81.95 & 96.83 & \underline{94.79} & 0.000012 \\
CTransPath & 92.92 & 89.03 & 97.31 & 94.61 & 89.73 & 92.21 & 98.90 & 85.34 & 80.77 & 95.79 & 91.66 & 0.000001 \\
\rowcolor{gray!40} SSMamba & 99.84 & 99.46 & 99.17 & \textbf{98.58} & \textbf{99.53} & \textbf{93.51} & \textbf{100.00} & 87.73 & \textbf{84.26} & \underline{97.68} & \textbf{95.98} & -\\
\bottomrule
\end{tabular}}
\end{subtable}

\vspace{1em}

\begin{subtable}[ht]{\textwidth}
\centering
\caption{Area Under Curve (AUC) Results Comparison of SSMamba with 11 SOTA Methods on 10 Pathology Datasets.}
\label{comparisonAUC}
\resizebox{\textwidth}{!}{
\begin{tabular}{lcccccccccccc} 
\toprule
\textbf{Method} & \textbf{LaC} & \textbf{NCT} & \textbf{PBC} & \textbf{pRCC} & \textbf{PAIP2019} & \textbf{CAM16} & \textbf{SIPa} & \textbf{MHIST} & \textbf{CRC} & \textbf{Ost} & \textbf{Ave-AUC} & \textbf{p-value}\\
\midrule
ViT & 90.92 & 84.63 & 91.55 & 87.45 & 86.25 & 84.27 & 56.81 & 80.76 & 78.44 & 90.40 & 83.15 & 0.000002 \\
Swin & 92.46 & 89.38 & 95.67 & 88.99 & 87.33 & 78.26 & 81.92 & 80.54 & 79.41 & 92.44 & 86.64 & 0.000003 \\
MAE & 92.67 & 83.19 & 93.12 & 90.74 & 86.62 & \underline{88.89} & 71.60 & 81.01 & 78.39 & 91.59 & 85.78 & 0.000002 \\
VMamba & 92.44 & 85.28 & 94.03 & 90.00 & 88.42 & 86.38 & 83.80 & 81.47 & 81.27 & 90.71 & 87.38 & 0.000001 \\
UNI & 96.08 & 91.14 & 96.76 & 88.08 & 95.51 & 79.06 & 64.79 & 84.38 & 85.45 & 95.27 & 87.65 & 0.000005 \\
UNI2 & 96.70 & 91.54 & 97.32 & 88.52 & 95.47 & 80.43 & 66.40 & 85.12 & \underline{86.18} & 96.17 & 88.38 & 0.000003 \\
MUSK & 93.64 & 93.19 & 98.54 & 91.48 & 94.71 & 83.39 & 85.19 & 84.58 & 80.90 & 95.27 & 90.09 & 0.000001 \\
CONCH & 92.52 & 91.60 & 98.16 & \textbf{95.27} & 83.85 & 74.36 & 91.37 & \underline{86.01} & 82.61 & 95.47 & 89.12 & 0.000001 \\
GigaPath & \underline{99.48} & \underline{99.36} & \textbf{99.51} & 92.92 & 94.25 & 72.87 & \underline{99.19} & \textbf{86.08} & 85.70 & \underline{96.60} & 92.60 & 0.000127 \\
Virchow2 & 98.07 & 99.28 & \underline{99.37} & 94.51 & \underline{97.05} & 77.40 & 98.47 & 85.66 & 85.60 & 96.03 & \underline{93.14} & 0.000015 \\
CTransPath & 94.47 & 90.75 & 97.09 & 91.95 & 88.43 & 83.89 & 98.90 & 82.72 & 82.94 & 93.57 & 90.47 & 0.000001 \\
\rowcolor{gray!40} SSMamba & \textbf{99.65} & \textbf{99.85} & \underline{99.37} & \underline{94.77} & \textbf{97.19} & \textbf{89.06} & \textbf{100.00} & 85.63 & \textbf{87.23} & \textbf{97.49} & \textbf{95.02} & - \\
\bottomrule
\end{tabular}}
\end{subtable}
\label{comparison}
\end{table*}

As demonstrated in Tables~\ref{comparisonF1},~\ref{comparisonAcc}, and~\ref{comparisonAUC}, comprehensive evaluations were conducted across 10 diverse pathological datasets to assess the performance of SSMamba against 11 SOTA models. The F1-score, accuracy, and Area Under the Curve (AUC) were adopted as core metrics to respectively evaluate the balance between precision and recall, classification reliability, and discriminative capability. SSMamba achieved outstanding results across all measures: an average F1-score of 95.56\%, an average accuracy of 95.98\%, and an average AUC of 95.02\%, outperforming the second-ranked model, Virchow2, by margins of 1.13\%, 1.19\%, and 1.88\%, respectively, thereby establishing a clear performance hierarchy. The model significantly surpassed traditional Transformer-based architectures (e.g., ViT, Swin) and even the Mamba-based VMamba (exceeding it by 9.36\% in F1-score), validating the efficacy of its optimized selective state space design. In detailed dataset-wise performance, SSMamba exhibited remarkable adaptability: it ranked first in all 10 tasks based on F1-score, including perfect 100.00\% scores on SIPa and top results on PBC, pRCC, PAIP2019, CAM16, and Ost; it also secured the highest accuracy on 5 datasets, including SIPa (100.00\%), pRCC, PAIP2019, CAM16, and CRC, while remaining highly competitive on the others; in terms of AUC, it achieved leading performance on 5 key datasets, including perfect 100.00\% on SIPa. SSMamba ranks first in 10 F1-score, 5 Accuracy, and 5 AUC tasks (e.g., 99.54\% F1-score on PBC, 99.53\% Accuracy on PAIP2019, 99.85\% AUC on NCT), and remains top-tier in the remaining tasks (e.g., 83.19\% F1-score on MHIST, only 0.66\% lower than GigaPath; 97.68\% Accuracy on Ost, 0.06\% lower than UNI2). Statistical analysis confirmed that all performance advantages of SSMamba over competing models were highly significant ($p<0.001$), effectively ruling out random variations and sampling bias. The core advantage of SSMamba stems from its optimized selective state space architecture, which enables linear-time modeling of long-sequence pathological features and dynamic information propagation. This design efficiently captures disease-specific diagnostic signals (e.g., cell morphology, tissue topology) while suppressing interference from normal tissues and mitigating domain shifts, thereby enhancing classification reliability and discriminative power across diverse pathological scenarios. 

\begin{figure}[!t]
\centering
\includegraphics[width=\linewidth]{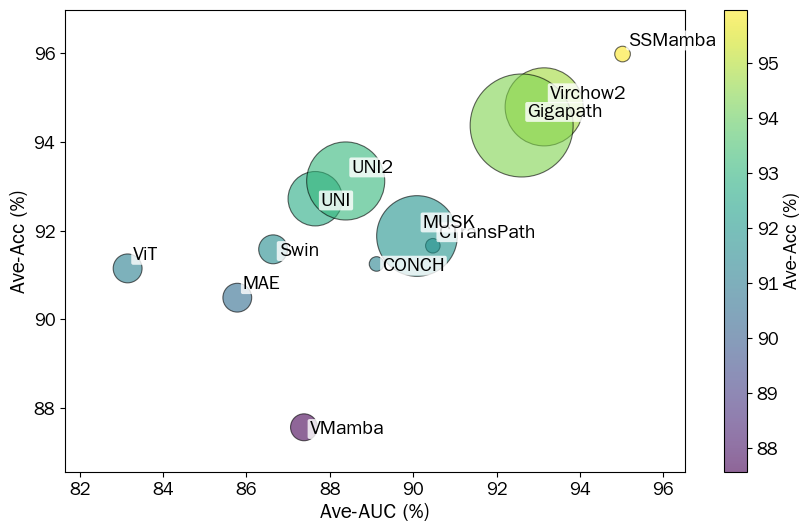}
\caption{Bubble plot of the effectiveness of SSMamba.}
\label{param}
\end{figure}

To intuitively and quantitatively demonstrate the trade-off between model performance and computational efficiency, we have added a new bubble chart (Fig.~\ref{param}). This chart plots Average Accuracy against Number of Parameters, where the size of the bubble indicates the number of parameters of the algorithm: the larger the bubble size, the more parameters the model has, and the smaller the bubble size, the fewer parameters it has. It can be clearly observed from the chart that compared with mainstream foundational models like Virchow2, our model achieves competitive or superior performance with a significantly reduced parameter count (25.3M parameters), fully demonstrating its advantages in computational efficiency and addressing the gap in resource consumption-related information.

\begin{table}[ht]
\centering
\caption{Performance Evaluation of Cross-Dataset Generalization.(The training data set is NCT. The test data sets correspond to MHIST and CRC, respectively.)}
\label{cross}
\resizebox{\linewidth}{!}{
\begin{tabular}{lcccccc}
\toprule
\multirow{2}{*}{\textbf{Method}} & \multicolumn{3}{c}{\textbf{MHIST}} & \multicolumn{3}{c}{\textbf{CRC}} \\
\cmidrule(r){2-4}\cmidrule(r){5-7}
& F1 & Acc & AUC & F1 & Acc & AUC \\
\midrule
UNI & 77.19 & 85.29 & 81.17 & 76.77 & 80.05 & 77.39\\
UNI2 & 78.00 & 85.97 & 82.03 & 77.06 & 80.10 & 78.03\\
MUSK & 82.00 & 86.83 & 84.77 & 77.31 & 71.13 & 78.13\\
CONCH & 80.26 & 84.09 & 84.01 & 83.79 & 78.66 & 78.52 \\
GigaPath & 82.31 & 86.59 & 84.29 & 83.43 & 80.44 & 83.63 \\
Virchow2 & 80.56 & 85.76 & 84.78 & 81.59 & 80.13 & 80.25 \\
\rowcolor{gray!40} SSMamba & 83.00 & 87.03 & 84.99 & 87.08 & 83.76 & 86.27 \\
\bottomrule
\end{tabular}}
\end{table}

\subsubsection{Evaluation of Cross-Dataset Generalization}
To validate the effectiveness of SSMamba in addressing cross-magnification domain shift and the limitation that fixed-scale pretraining fails to adapt to diverse clinical scenarios, we conduct cross-dataset generalization experiments. The model is trained on the NCT dataset and evaluated on two independent test datasets (MHIST and CRC), with results summarized in Table~\ref{cross}. On the MHIST dataset, SSMamba achieves balanced and superior performance across F1-score (83.00\%), accuracy (Acc, 87.03\%), and AUC (84.99\%). Compared with SOTA baselines, it outperforms GigaPath (F1=82.31\%, Acc=86.59\%, AUC=84.29\%) and MUSK (F1=82.00\%, Acc=86.83\%, AUC=84.77\%) in all metrics. This indicates that SSMamba can effectively capture discriminative structural features from histopathological images, which helps mitigate cross-magnification domain shift. For the CRC dataset, SSMamba maintains stable and leading performance, with F1-score of 87.08\%, Acc of 83.76\%, and AUC of 86.27\%. It outperforms the second-best baseline GigaPath by 3.65\% in F1-score, 3.32\% in Acc, and 2.64\% in AUC. In contrast, other baselines show inconsistent performance across datasets: MUSK achieves relatively high F1 on MHIST but drops to 71.13\% in Acc on CRC, reflecting poor adaptability to diverse scenarios.       SSMamba’s consistent performance across two datasets verifies its ability to overcome the limitation of fixed-scale pretraining. The above results demonstrate that SSMamba’s MAMIM balances the retention of discriminative contextual cues and the enhancement of domain-invariant feature learning through a targeted MIM paradigm. Meanwhile, SSMamba’s hierarchical SSM-based encoder contributes to integrating multi-scale features and modeling long-range contexts, thereby enhancing its robustness to cross-magnification domain shift and adaptability to diverse clinical scenarios.

\begin{table*}[htbp]
\centering
\caption{The Performance Comparison with 8 SOTA methods on 6 Downstream Datasets. \textbf{Bold}: the best result. \underline{Underline}: the second best result.}
\label{tab:WSI}
\resizebox{\textwidth}{!}{
\begin{tabular}{lcccccccccccccccccc}
\toprule
\multirow{3}{*}{\textbf{Method}} & \multicolumn{3}{c}{\textbf{PANDA}} & \multicolumn{3}{c}{\textbf{TCGA-ESCA}} & \multicolumn{3}{c}{\textbf{TCGA-BRCA}} & \multicolumn{3}{c}{\textbf{TCGA-CESC}} & \multicolumn{3}{c}{\textbf{TCGA-LGG}} & \multicolumn{3}{c}{\textbf{TCGA-BLCA}} \\
& \multicolumn{3}{c}{\textbf{I-Grade}} & \multicolumn{3}{c}{\textbf{Grade}} & \multicolumn{3}{c}{\textbf{IHC-HER2}} & \multicolumn{3}{c}{\textbf{LymInv}} & \multicolumn{3}{c}{\textbf{T-Stage}} & \multicolumn{3}{c}{\textbf{OS}} \\
\cmidrule(lr){2-4} \cmidrule(lr){5-7} \cmidrule(lr){8-10} \cmidrule(lr){11-13} \cmidrule(lr){14-16} \cmidrule(lr){17-19}
& AUC & F1 & Acc & AUC & F1 & Acc & AUC & F1 & Acc & AUC & F1 & Acc & AUC & F1 & Acc & \multicolumn{3}{c}{Corr}\\
\midrule
SlideAve & 89.95 & 60.76 & 66.93 & 57.37 & 15.22 & 38.89 & 55.99 & 23.28 & 59.02 & 49.45 & 44.14 & 44.44 & 70.85 & \textbf{70.47} & \underline{69.47} & \multicolumn{3}{c}{56.00} \\
SlideMax & 94.88 & 70.52 & 76.68 & 55.18 & 14.89 & 38.89 & \underline{64.32} & \underline{27.92} & 57.92 & 50.55 & 51.59 & 51.85 & 62.82 & 57.05 & 55.80 & \multicolumn{3}{c}{51.08} \\
ABMIL & 94.56 & 69.70 & 75.93 & 54.33 & 21.79 & \underline{41.67} & 59.19 & 25.60 & 60.11 & 56.04 & 55.49 & 55.56 & 58.35 & 51.01 & 50.93 & \multicolumn{3}{c}{\textbf{61.20}} \\
CLAM & 94.81 & 71.68 & 77.48 & 61.05 & 25.34 & \textbf{47.22} & 58.52 & 27.34 & \underline{62.84} & 58.24 & 55.00 & 55.56 & 58.63 & 58.39 & 57.70 & \multicolumn{3}{c}{46.05} \\
DSMIL & 94.88 & 71.39 & 77.11 & 55.59 & 20.33 & \underline{41.67} & 62.19 & 22.46 & 60.11 & \underline{60.44} & 48.08 & 48.15 & 64.06 & 59.06 & 56.81 & \multicolumn{3}{c}{55.52} \\
GigaPath & 94.16 & 69.27 & 74.00 & 64.52 & \textbf{31.59} & \underline{41.67} & 54.37 & 22.14 & 49.73 & 51.65 & 55.49 & 55.56 & 64.04 & 61.74 & 60.89 & \multicolumn{3}{c}{50.37} \\
S4MIL & \textbf{95.78} & \underline{74.40} & \underline{79.79} & 66.64 & 25.12 & \textbf{47.22} & 60.02 & \textbf{28.66} & 61.20 & 49.45 & 50.00 & 55.56 & 67.53 & 60.40 & 52.74 & \multicolumn{3}{c}{55.10} \\
MambaMIL & 92.74 & 62.90 & 71.36 & 70.63 & 20.38 & \underline{41.67} & 62.56 & 22.74 & 61.20 & 52.20 & \underline{58.35} & \underline{59.26} & \underline{72.87} & 69.80 & 68.69 & \multicolumn{3}{c}{57.86} \\
\rowcolor{gray!40}SSMambaMIL & \underline{95.40} & \textbf{75.31} & \textbf{79.91} & \textbf{71.83} & \underline{27.35} & \textbf{47.22} & \textbf{64.71} & 27.74 & \textbf{63.39} & \textbf{64.56} & \textbf{62.96} & \textbf{62.90} & \textbf{77.41} & \underline{69.93} & \textbf{71.81} & \multicolumn{3}{c}{\underline{59.28}} \\
\bottomrule
\end{tabular}}
\end{table*}

\subsubsection{Downstream Tasks of WSI Pathological Image}
As shown in Table~\ref{tab:WSI}, SSMambaMIL delivers superior overall performance across six downstream tasks on six datasets, achieving the best results (bold) in 12 out of 17 metrics and the second-best (underline) in 4 metrics. It outperforms 8 SOTA methods, including MIL-based approaches (ABMIL, CLAM, DSMIL) and SSM-based models (S4MIL, MambaMIL). On the PANDA dataset’s I-Grade task, SSMambaMIL ranks first in F1-score and Accuracy, which are 0.91\% and 0.12\% higher than the second-ranked S4MIL respectively. In terms of AUC, S4MIL takes the lead, while SSMambaMIL ranks second with a narrow gap. 

For the TCGA-ESCA Grade task, SSMambaMIL secures the top position in AUC and Acc. Its AUC is 1.20\% higher than MambaMIL, the second-ranked method in this metric. In Acc, it ties with CLAM and S4MIL for the first place. In F1-score, GigaPath ranks first, and SSMambaMIL takes the second place, which is higher than most competing methods such as CLAM and S4MIL. In the IHC-HER2 task of TCGA-BRCA, SSMambaMIL performs best in AUC and Acc. Its AUC is 0.39\% higher than SlideMax, the second-ranked method, and its Acc is 0.55\% higher than the second-best CLAM. S4MIL ranks first in F1-score (28.66\%), with SSMambaMIL following closely in the second place (27.74\%). Notably, SSMambaMIL achieves a clean sweep of the first place in all three metrics on the LymInv task of TCGA-CESC. Specifically, its AUC is 4.12\% higher than DSMIL, the second-ranked method, and its F1-score is 4.61\% higher than the second-place MambaMIL. On the T-Stage task of TCGA-LGG, SSMambaMIL ranks first in both AUC and Acc.  Its AUC is 4.54\% higher than the second-ranked MambaMIL, and its Acc is 3.12\% higher than SlideAve, which takes the second place.  For the OS regression task of TCGA-BLCA, ABMIL ranks first in Correlation, while SSMambaMIL ranks second, outperforming MambaMIL and SlideAve. These results confirm that SSMambaMIL’s advanced sequence modeling capability can effectively capture discriminative features in WSIs.

\subsection{Visualization Analysis}
\begin{figure*}[!t]
\centering
\includegraphics[width=\linewidth]{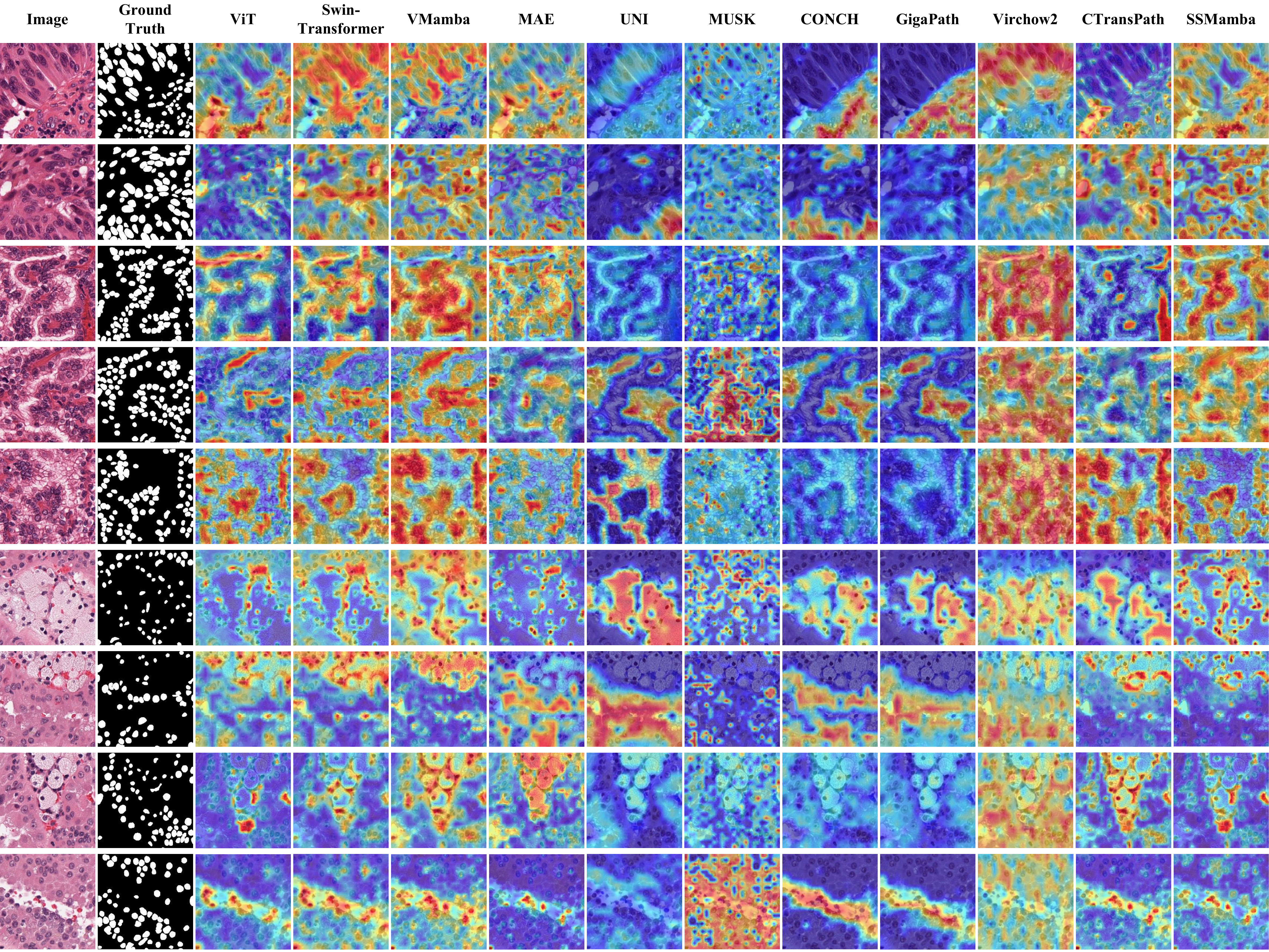}
\caption{Grad-CAM Visualizations of Feature Representations on ROI Pathological Image.}
\label{fig: cam}
\end{figure*}
Figure~\ref{fig: cam} presents a qualitative comparison of feature representations produced by different vision models using Gradient Class Activation Maps (Grad-CAM) on Region of Interest (ROI) patches from Warwick dataset~\cite{Warwick}. We use Warwick for this qualitative analysis because it shares the same pathology type as NCT (H\&E-stained colorectal adenocarcinoma), while additionally providing pixel-level annotations (e.g., segmentation masks) that serve as a convenient visual reference for assessing whether highlighted regions align with disease-relevant structures. Therefore, all models are trained only on the NCT dataset, and Warwick is introduced solely as an independent cohort for visualization. This setting allows to examine whether representations learned from NCT transfer to a different dataset with consistent morphology, and make a clearer side-by-side comparison between SSMamba and competing methods in terms of where and how they attend. The visualization is structured as follows: the first column displays the original input patches; the second column shows the corresponding ground-truth annotations; subsequent columns present the Grad-CAM generated by each evaluated algorithm. A critical analysis of these visualizations reveals distinct characteristics, along with potential strengths and limitations, across models:

\textbf{ViT} excels at capturing global architecture but tends to over-homogenize fine-grained cellular structures, which may blur critical boundaries (e.g., nuclear-cytoplasmic interfaces) in some cases; \textbf{Swin} enhances local localization capabilities but may struggle to fully integrate long-range dependencies, leading to potential fragmentation of attention between related structures; \textbf{VMamba} demonstrates efficiency advantages but may generate artifactual "striping" due to fixed unidirectional scanning; \textbf{MAE} is effective at reconstructing low-level statistics but may over-smooth or fragment nuclear and cellular boundaries to some extent; \textbf{UNI} tends to neglect hierarchical structures due to rigid tokenization, which may result in the loss of subcellular details (manifested as minimal regions of interest, mostly blue) in relevant scenarios; \textbf{MUSK} may dilute micro-scale features, potentially leading to misfocus on negative samples in sparse tumor regions; \textbf{CONCH} may introduce noise from heuristic labels, with attention possibly scattered over non-abnormal areas in some cases; \textbf{Gigapath} can induce spatial fragmentation at tile boundaries under certain circumstances, which may sever continuous patterns (e.g., microvascular invasion); \textbf{Virchow2} may struggle to resolve 3D relationships effectively and is relatively vulnerable to staining variations, often resulting in large indistinct regions; \textbf{CTransPath} may dilute discriminative features in some scenarios, potentially misfocusing on negative samples in partial images.

In contrast, \textbf{SSMamba} integrates strengths of these models while mitigating their limitations. It retains ViT’s grasp of global tissue architecture without sacrificing fine-grained cellular details, ensuring sharp delineation of critical boundaries like nuclear-cytoplasmic interfaces. By seamlessly bridging local feature precision (as in Swin) and long-range biological dependencies, it avoids fragmented attention, enabling holistic assessment of tissue structures. Free from VMamba’s "striping" artifacts and MAE’s over-smoothed boundaries, SSMamba preserves subcellular nuances essential for grading subtle malignancies, while its flexible tokenization outperforms UNI’s rigidity. Unlike MUSK’s diluted micro-scale features or CONCH’s scattered attention, it hones in on diagnostically critical regions, and it avoids Gigapath’s spatial fragmentation and Virchow2’s vulnerability to staining variations. By prioritizing clinically consequential features over irrelevant areas (unlike CTransPath), SSMamba sets a new standard for robust, context-aware pathological image analysis.
\section{Discussion}
\label{discussion}
To comprehensively evaluate the effectiveness of each proposed component, we conducted a series of ablation studies on 6 benchmark pathology image datasets. This section validates SSMamba's three core components (MAMIM, DMS, LPR) via 3 subsections. Each subsection combines quantitative and qualitative analyses on public datasets to verify their efficacy in addressing pathology model limitations.

\begin{figure*}[!t]
\centering
\includegraphics[width=\linewidth]{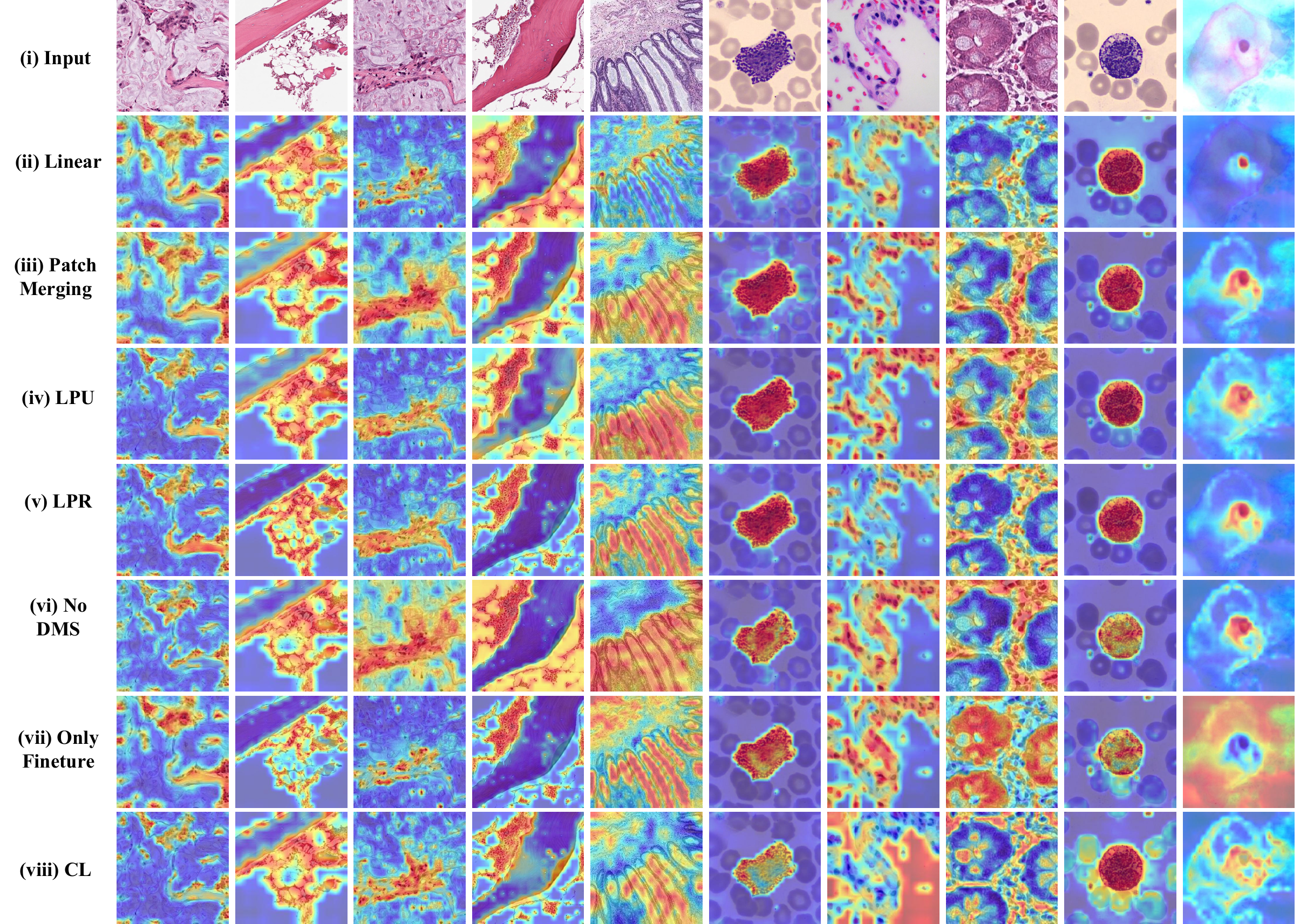}
\caption{Grad-CAM visualization of the influence of different modules on the feature representation of ROI pathology images. ((i) original image; (ii) SSMamba with Linear projection; (iii) SSMamba with Patch Merge; (iv) SSMamba with Local Perception Unit (LPU); (v) SSMamba with LPR (our final model); (vi) SSMamba using traditional Mamba modules; (vii) SSMamba without pre-training; (viii) SSMamba pre-trained in CL.)}
\label{fig: diss}
\end{figure*}

\begin{table*}[t]
\centering
\caption{Performance Evaluation of MAMIM Pretraining Strategy. (The table compares the performance of different pre-training strategies across 6 pathology datasets.)}
\label{tab:MAMIM}
\resizebox{\textwidth}{!}{
\begin{tabular}{lcccccccccccccccccc}
\toprule
\textbf{Pre-training} & \multicolumn{3}{c}{\textbf{NCT}} & \multicolumn{3}{c}{\textbf{PBC}} & \multicolumn{3}{c}{\textbf{pRCC}} & \multicolumn{3}{c}{\textbf{CAM16}} & \multicolumn{3}{c}{\textbf{MHIST}} & \multicolumn{3}{c}{\textbf{CRC}} \\
\cmidrule(lr){2-4} \cmidrule(lr){5-7} \cmidrule(lr){8-10} \cmidrule(lr){11-13} \cmidrule(lr){14-16} \cmidrule(lr){17-19}
 & \textbf{F1} & \textbf{Acc} & \textbf{AUC} & \textbf{F1} & \textbf{Acc} & \textbf{AUC} & \textbf{F1} & \textbf{Acc} & \textbf{AUC} & \textbf{F1} & \textbf{Acc} & \textbf{AUC} & \textbf{F1} & \textbf{Acc} & \textbf{AUC} & \textbf{F1} & \textbf{Acc} & \textbf{AUC} \\
\midrule
None & 90.07 & 90.50 & 90.50 & 95.33 & 95.08 & 95.08 & 94.07 & 94.95 & 94.95 & 87.44 & 88.65 & 88.65 & 81.48 & 83.69 & 84.48 & 87.55 & 83.49 & 85.38 \\
CL & 96.90 & 96.99 & 96.99 & 97.70 & 97.98 & 97.98 & 96.77 & 97.91 & 97.91 & 90.50 & 90.35 & 90.35 & 82.12 & 84.07 & 84.92 & 88.71 & 85.11 & 86.21 \\
MAE & 97.00 & 97.03 & 97.03 & 98.07 & 97.81 & 97.81 & 96.43 & 97.91 & 97.91 & 88.85 & 90.33 & 90.33 & 80.70 & 82.78 & 83.47 & 87.29 & 82.83 & 84.69 \\
\rowcolor{gray!20}
MAMIM & \textbf{99.38} & \textbf{99.46} & \textbf{99.85} & \textbf{99.54} & \textbf{99.17} & \textbf{99.37} & \textbf{97.87} & \textbf{98.58} & 94.77 & \textbf{91.87} & \textbf{93.51} & \textbf{89.06} & \textbf{83.19} & \textbf{87.73} & \textbf{85.63} & \textbf{89.80} & \textbf{84.26} & \textbf{87.23} \\
\bottomrule
\end{tabular}}
\end{table*}

\subsection{Effectiveness of the MAMIM}
\subsubsection{Analysis of the MAMIM's Quantitative Comparison Results}
To mitigate domain shift in ROI data, MAMIM adopts the DMS module as a dedicated decoder for reconstructing partially masked pathological ROI inputs in SSMamba. Table~\ref{tab:MAMIM} validates MAMIM against three baselines: no pretraining, CL-based pretraining (CTransPath), and MAE pretraining. For NCT, it delivers Acc 99.46\%, F1-score 99.38\%, and a notable AUC of 99.85\%, outperforming all baselines by resolving nuclear morphology distortions from multi-center variations. On PBC, its top F1-score of 99.54\% represents improvements of 4.21\%, 1.84\%, and 1.47\% over the three baselines, overcoming CL’s limitation in handling cross-domain intensity variations via precise masked leukocyte reconstruction. For pRCC, MAMIM’s F1-score of 97.87\% (3.80\% higher than no pretraining, 1.10\% higher than CL, 1.44\% higher than MAE) preserves glandular continuity and reduces domain-induced feature fragmentation. For CAM16, it achieves Acc improvements of 4.86\%, 3.16\%, and 3.18\% over the three baselines, enhancing cross-domain transferability critical for subtle lesion detection. Across MHIST and CRC, MAMIM maintains leading performance: top metrics on MHIST, F1-score (89.80\%) and AUC (87.23\%) on CRC. Overall, quantitative results across 6 datasets demonstrate MAMIM’s consistent superiority over all baselines in F1-score, Acc, and AUC, validating its domain-invariant pretraining mechanism for enhanced cross-institutional generalization.

\subsubsection{The MAMIM's Grad-CAM Visual Result Analysis}
Analysis of Fig.~\ref{fig: diss}(v, vii, viii) provides insights into the feature representation strengths of our proposed MAMIM. Notably, MAMIM draws on the masking-based learning paradigm but differs fundamentally from Masked Autoencoders (MAE): instead of relying on Vision Transformers (ViT) as MAE does, MAMIM integrates masking strategies with VMamba to optimize feature learning for pathological images. Therefore, MAMIM performs masking-based self-supervised learning (SSL) directly on target ROI data with consistent magnification and field-of-view to downstream tasks, leveraging a 75\% high masking ratio to infer occluded tissue regions. This design inherently shifts attention from artifacts to diagnostically relevant structural signals (e.g., cell morphology, tissue topology) while alleviating cross-magnification domain shift—advantages observed across diverse pathological datasets. Unpretrained SSMamba consistently fixates on non-diagnostic artifacts (e.g., uneven staining, low-magnification blur, impurities) due to the absence of guidance for prioritizing clinically meaningful signals. CL-pretrained SSMamba suffers from compromised feature discrimination and generalization, primarily caused by cross-magnification domain shift induced by mismatched FOV between pretraining and downstream tasks. This mismatch leads to issues such as misidentification of benign/malignant tissues, misattribution to background artifacts, and poor cross-magnification adaptability. However, MAMIM effectively suppresses magnification-specific noise and irrelevant artifacts.

\begin{figure*}[!t]
\centering
\includegraphics[width=\linewidth]{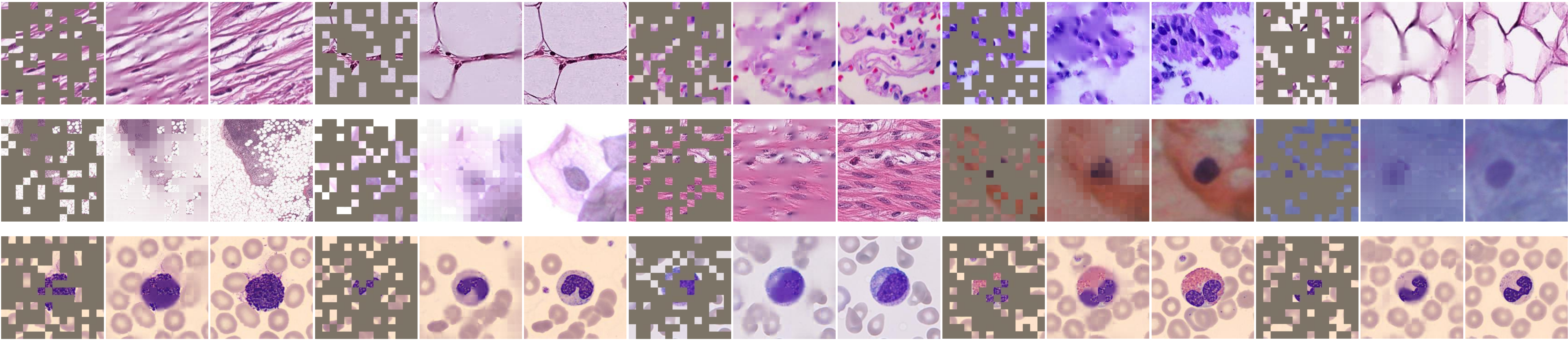}
\caption{Masked Reconstruction Samples on ROI Image (masking ratio: 75\%). Left: Masked input; Middle: SSMamba reconstruction; Right: Ground truth.}
\label{vision}
\end{figure*}

\subsubsection{Masked Reconstruction Performance Analysis}
To further validate SSMamba's robustness against domain shifts, we assessed its masked reconstruction performance across datasets with substantial staining and processing variations. As illustrated in Fig.~\ref{vision}, SSMamba retains exceptional diagnostic fidelity across all scenarios: NCT: Preserves subtle nuclear chromatin patterns across 5+ hospital sources. LaC: Reconstructs consistent glandular architectures resilient to H\&E staining batch variations. PBC: Maintains precise boundaries between lymphocytes and myelocytes amid fixation artifacts. pRCC: Preserves papillary core structures and clear cell cytoplasmic integrity across 12 institutional staining protocols. SIPa: Accurately reconstructs diagnostic nuclear chromocenters and cytoplasmic keratinization despite 3× staining concentration variations and liquid-based preparation smearing. PAIP2019: Maintains sharp tumor margins and microvascular invasion traces amid heterogeneous necrosis. CAM16: Preserves single tumor cell distribution and lymphocyte infiltration topology across 8 scanner types. MHIST: Retains nuclear pleomorphism and ductal and lobular topology across magnifications, mitigating cross-magnification domain shift via field-of-view matched pretraining. CRC: Preserves glandular disorganization and invasive fronts across staining variations. Ost: Maintains pleomorphic tumor cell morphology and abnormal bone matrix infiltration amid calcification artifacts, resisting cross-magnification domain shift via magnification-matched SSL. Overall, SSMamba’s reconstructions (middle column) align closely with the ground truth (right column), confirming its ability to learn domain-invariant representations.

\subsection{Effectiveness of the DMS Module}
\subsubsection{Analysis of the DMS’s Quantitative Comparison Results}
\begin{table*}[htbp]
\centering
\caption{Performance Evaluation of the DMS Module. (This table presents the performance evaluation of the DMS module on 6 selected pathology datasets (NCT, PBC, pRCC, CAM16, MHIST, and CRC). ``o/DMS'' indicates without the DMS module, while ``w/DMS'' indicates with the DMS module.)}
\label{tab: DMS}
\resizebox{\textwidth}{!}{
\begin{tabular}{lcccccccccccccccccc}
\toprule
\textbf{Method} & \multicolumn{3}{c}{\textbf{NCT}} & \multicolumn{3}{c}{\textbf{PBC}} & \multicolumn{3}{c}{\textbf{pRCC}} & \multicolumn{3}{c}{\textbf{CAM16}} & \multicolumn{3}{c}{\textbf{MHIST}} & \multicolumn{3}{c}{\textbf{CRC}} \\
\cmidrule(lr){2-4} \cmidrule(lr){5-7} \cmidrule(lr){8-10} \cmidrule(lr){11-13} \cmidrule(lr){14-16} \cmidrule(lr){17-19}
 & \textbf{F1} & \textbf{Acc} & \textbf{AUC} & \textbf{F1} & \textbf{Acc} & \textbf{AUC} & \textbf{F1} & \textbf{Acc} & \textbf{AUC} & \textbf{F1} & \textbf{Acc} & \textbf{AUC} & \textbf{F1} & \textbf{Acc} & \textbf{AUC} & \textbf{F1} & \textbf{Acc} & \textbf{AUC} \\
\midrule
VMamba o/DMS & 90.80 & 91.57 & 85.28 & 87.19 & 85.33 & 94.03 & 84.20 & 86.39 & 90.00 & 84.40 & 86.69 & 86.38 & 76.87 & 85.33 & 81.47 & 82.84 & 79.07 & 81.27 \\
VMamba w/DMS & 92.47 & 93.87 & 88.02 & 89.47 & 87.76 & 94.19 & 86.15 & 88.07 & 91.33 & 86.36 & 88.66 & 86.50 & 77.50 & 86.00 & 82.74 & 84.07 & 79.97 & 82.90 \\
\midrule
SSMamba o/DMS & 95.91 & 97.30 & 98.79 & 97.75 & 97.80 & 97.83 & 94.57 & 96.01 & 90.88 & 87.88 & 88.96 & 88.36 & 81.58 & 85.45 & 83.58 & 87.33 & 82.03 & 85.40 \\
\rowcolor{gray!20} SSMamba w/DMS & \textbf{99.38} & \textbf{99.46} & \textbf{99.85} & \textbf{99.54} & \textbf{99.17} & \textbf{99.37} & \textbf{97.87} & \textbf{98.58} & \textbf{94.77} & \textbf{91.87} & \textbf{93.51} & \textbf{89.06} & \textbf{83.19} & \textbf{87.73} & \textbf{85.63} & \textbf{89.80} & \textbf{84.26} & \textbf{87.23} \\
\bottomrule
\end{tabular}}
\end{table*}

\begin{table*}[htbp]
\centering
\caption{Performance Evaluation of the LPR Module on Selected Datasets. (This table presents the performance evaluation of the LPR module on 6 selected pathology datasets (NCT, PBC, pRCC, CAM16, MHIST, and CRC).)}
\label{tab: LPR}
\resizebox{\textwidth}{!}{
\begin{tabular}{lcccccccccccccccccc}
\toprule
\textbf{Method} & \multicolumn{3}{c}{\textbf{NCT}} & \multicolumn{3}{c}{\textbf{PBC}} & \multicolumn{3}{c}{\textbf{pRCC}} & \multicolumn{3}{c}{\textbf{CAM16}} & \multicolumn{3}{c}{\textbf{MHIST}} & \multicolumn{3}{c}{\textbf{CRC}} \\
\cmidrule(lr){2-4} \cmidrule(lr){5-7} \cmidrule(lr){8-10} \cmidrule(lr){11-13} \cmidrule(lr){14-16} \cmidrule(lr){17-19}
 & \textbf{F1} & \textbf{Acc} & \textbf{AUC} & \textbf{F1} & \textbf{Acc} & \textbf{AUC} & \textbf{F1} & \textbf{Acc} & \textbf{AUC} & \textbf{F1} & \textbf{Acc} & \textbf{AUC} & \textbf{F1} & \textbf{Acc} & \textbf{AUC} & \textbf{F1} & \textbf{Acc} & \textbf{AUC} \\
\midrule
Linear & 98.86 & 98.98 & 98.98 & 98.17 & 98.22 & 98.22 & 96.67 & 97.33 & 97.33 & 87.92 & 90.37 & 90.37 & 81.49 & 85.17 & 84.44 & 87.47 & 82.79 & 85.71 \\
Patch Merge & 99.02 & 99.31 & 99.31 & 99.11 & 98.98 & 98.98 & 97.19 & 98.00 & 98.00 & 90.06 & 91.88 & 91.88 & 81.88 & 86.41 & 84.90 & 88.08 & 82.96 & 86.63 \\
LPU & 98.90 & 99.20 & 99.20 & 99.04 & 98.70 & 98.70 & 96.97 & 97.97 & 97.97 & 89.17 & 90.69 & 90.69 & 82.07 & 87.05 & 85.18 & 88.66 & 83.46 & 86.83 \\
\rowcolor{gray!20} \textbf{LPR} & \textbf{99.38} & \textbf{99.46} & \textbf{99.46} & \textbf{99.54} & \textbf{99.17} & \textbf{99.17} & \textbf{97.87} & \textbf{98.58} & \textbf{98.58} & \textbf{91.87} & \textbf{93.51} & \textbf{93.51} & \textbf{83.19} & \textbf{87.73} & \textbf{85.63} & \textbf{89.80} & \textbf{84.26} & \textbf{87.23} \\
\bottomrule
\end{tabular}}
\end{table*}

To further evaluate the efficacy of the proposed DMS module, we conduct comparative experiments among four configurations: the original VMamba (VMamba o/DMS), VMamba with our DMS module replacing the original VMamba module (VMamba w/DMS), SSMamba with the original VMamba module (SSMamba o/DMS), and SSMamba with the DMS module (SSMamba w/DMS). As shown in Table~\ref{tab: DMS}, the DMS-enhanced architecture outperforms all baselines across 6 datasets. On the NCT dataset, SSMamba w/DMS reaches 99.46\% accuracy. DMS’s multi-scale directional kernels aggregate nuclear features, preserve local-global histological context, and address the original decoder’s deficiency in fine-grained spatial feature extraction. On the PBC dataset, SSMamba w/DMS improves Acc, F1-score, and AUC by 1.37\%, 1.79\%, and 1.54\% over SSMamba o/DMS, respectively. DMS’s direction-aware fusion integrates erythrocyte and leukocyte features, eliminates edge artifacts from staining variations, and enhances target cell identification. On the pRCC dataset, SSMamba w/DMS achieves 98.58\% accuracy, 97.87\% F1-score, and 94.77\% AUC. DMS’s trajectory-aligned convolutions model glandular continuity, resolve tumor-stroma fragmentation, and improve tissue differentiation. On the CAM16 dataset, SSMamba w/DMS improves Acc, F1-score, and AUC by 4.55\%, 3.99\%, and 0.70\% over SSMamba o/DMS. DMS’s hierarchical directional receptive fields retain cellular and tissue-level features, mitigating downsampling losses and enhancing subtle lesion discriminability. For MHIST and CRC, SSMamba w/DMS maintains leading performance. In summary, DMS enhances the model’s ability to capture fine-grained structural features and spatial relationships. Consistent improvements across 6 datasets validate the effectiveness and generalization of DMS when integrated with SSMamba.

\subsubsection{The DMS’s Grad-CAM Visual Result Analysis}
Analysis of Fig.~\ref{fig: diss}(v, vi) reveals consistent feature activation visualization patterns across diverse datasets, underscoring the functional contribution of our proposed DMS module. Specifically, without the DMS module incorporated (SSMamba o/DMS), the class activation maps are characterized by prominent artifacts. The model exhibits a propensity for global attention coverage while lacking accurate localization of local structures, which manifests as coarse-grained and indistinct cell boundaries that fail to resolve fine-grained structural details. Furthermore, spurious background activation is observed in certain images, where non-diagnostic background regions are inappropriately accentuated at the expense of core cellular and tissue regions of interest. In contrast, the class activation maps of SSMamba integrated with the DMS module (SSMamba w/DMS) demonstrate remarkable improvements in visualization fidelity. Artifacts are markedly mitigated, and the model shifts from a pattern of exclusive global attention to a balanced global-local attention framework. Cell boundaries exhibit enhanced sharpness and clarity, with concomitant precise delineation of fine-grained structural details. Furthermore, background confounding signals are efficiently suppressed, ensuring that the model’s attention is selectively directed toward target cellular and tissue regions rather than non-diagnostic background compartments.

\subsection{Effectiveness of the LPR Module}
\subsubsection{Analysis of the LPR’s Quantitative Comparison Results}
To further enhance spatial representation, we extend the proposed SSMamba framework and compare our LPR module with three commonly employed embedding methods: (1) Linear projection (ViT-style), (2) Patch Merge (Swin-style), and (3) Locality-Preserving Unit (LPU). Table~\ref{tab: LPR} presents the performance of the LPR module against three baseline embedding methods (Linear PE, Patch Merge, LPU) across 6 pathological datasets, with F1-score, Accuracy (Acc), and AUC as evaluation metrics. Experimental results confirm that LPR outperforms all baselines consistently across all metrics, verifying its effectiveness for pathology-specific feature embedding. Specifically, Table~\ref{tab: LPR} shows LPR achieves the highest F1-scores on 6 datasets: 99.54\% on PBC (4.21\% improvement over Linear PE). For Acc and AUC, LPR maintains SOTA performance, including perfect scores on CAM16 (Acc: 93.51\%, 3.14\% higher than Patch Merge; AUC: 93.51\%, 2.63\% above LPU). Even on MHIST and CRC where baseline performance is already competitive, LPR still delivers marginal but consistent improvements, e.g., 89.80\% F1-score on CRC (1.14\% higher than LPU). These consistent performance gains validate that LPR’s integration of depthwise convolution and residual fusion effectively enhances translation invariance and robustness to staining artifacts. Its lightweight architecture further ensures efficient and reliable feature extraction across diverse pathological scenarios.

\subsubsection{The LPR’s Grad-CAM Visual Result Analysis}
Analysis of Fig.~\ref{fig: diss} (ii-v) demonstrates distinct visualization patterns across different modules, highlighting the advantages of our designed LPR module. For linear projection, the generated class activation maps exhibit notable positional sensitivity and prominent artifacts. Identical features across different spatial locations show inconsistent activation intensities, with erroneous attention allocation to non-diagnostic regions such as areas with staining diffusion, tissue folds, and ductal structures. Additionally, hierarchical downsampling via patch merge induces excessive compression of spatial information, which exacerbates staining-related artifacts. This manifests as blurred structural boundaries, degraded fine-grained details, obscured key morphological features, and weakened attention responses in heterogeneously stained regions and tissue interfaces. In comparison, the class activation maps of the LPU module show improved edge delineation relative to linear projection but remain limited in visualization stability. The maps lack robustness to staining variations, with excessive attention focused on regions affected by residual dye. Moreover, long-range spatial correlation modeling is inadequate, leading to spurious attention to background areas and failure to capture the spatial relationships between target tissues and adjacent structures. In contrast, the class activation maps of SSMamba integrated with the proposed LPR module exhibit remarkable optimization across diverse datasets. The module effectively mitigates positional sensitivity, resulting in stable activation patterns for identical features regardless of spatial location. Visual artifacts are significantly reduced, with clear structural boundaries, well-preserved fine-grained morphological details, and balanced attention allocation in heterogeneously stained regions. Background interference is minimized, and the spatial relationships between target tissues and adjacent structures are accurately reflected in the activation maps.
\section{Conclusion}
\label{conclusion}
In this work, we pinpoint three key limitations in existing pathology FMs: (1) sensitivity to translations and rotations, (2) inefficient local–global feature integration, and (3) susceptibility to scanner‑ and stain‑induced domain shifts. To address them, we present SSMamba, a task‑specific, two‑stage framework that couples in‑domain SSL pretraining (MAMIM) on ROI data with lightweight supervised fine‑tuning. Its hybrid state‑space backbone incorporates a Local Perception Residual (LPR) module for translation and rotation invariance and a Directional Multi‑scale (DMS) module for hierarchical feature fusion. Tested on 10 public ROI datasets, SSMamba surpasses 6 leading pathology FMs (e.g., +1.83 F1 on CAM16 over CTransPath), demonstrating that pathology‑aware architectures and targeted in‑domain SSL achieve superior accuracy and robustness without billion‑scale pretraining. Meanwhile, SSMambaMIL delivers superior overall performance across 6 downstream tasks on 6 WSI datasets. It outperforms 8 SOTA methods, including MIL-based approaches (ABMIL, CLAM, DSMIL) and SSM-based models (S4MIL, MambaMIL).  

Despite these promising results, the current scope of our framework has a notable limitation: our validation and optimization are exclusively focused on pathological image classification tasks (both ROI-level and WSI-level).  Future work will extend the SSMamba framework to other pathology-relevant tasks, such as semantic segmentation of tissue regions and mitosis detection.
\section{Acknowledgments}
\subsection{Declaration of competing interest}
The authors declare that they have no known competing financial interests or personal relationships that could have appeared to influence the work reported in this paper.

\subsection{Data availability}
The authors use public datasets. The public datasets can be downloaded from the corresponding official websites.
\bibliography{Sections/References}

@STRING{CVPR = "Proc. IEEE Conf. Comput. Vis. Pattern Recognit."}

@STRING{AAAI = "Proc. AAAI Conf. Artif. Intell."}

@inproceedings{dsmil,
  title={Dual-stream multiple instance learning network for whole slide image classification with self-supervised contrastive learning},
  author={Li, Bin and Li, Yin and Eliceiri, Kevin W},
  booktitle={Proceedings of the IEEE/CVF conference on computer vision and pattern recognition},
  pages={14318--14328},
  year={2021}
}

@inproceedings{selfmedmae,
  title={Self pre-training with masked autoencoders for medical image classification and segmentation},
  author={Zhou, Lei and Liu, Huidong and Bae, Joseph and He, Junjun and Samaras, Dimitris and Prasanna, Prateek},
  booktitle={2023 IEEE 20th International Symposium on Biomedical Imaging (ISBI)},
  pages={1--6},
  year={2023},
  organization={IEEE}
}

@inproceedings{lerousseau2020weakly,
  title={Weakly supervised multiple instance learning histopathological tumor segmentation},
  author={Lerousseau, Marvin and Vakalopoulou, Maria and Classe, Marion and Adam, Julien and Battistella, Enzo and Carr{\'e}, Alexandre and Estienne, Th{\'e}o and Henry, Th{\'e}ophraste and Deutsch, Eric and Paragios, Nikos},
  booktitle={Medical Image Computing and Computer Assisted Intervention--MICCAI 2020: 23rd International Conference, Lima, Peru, October 4--8, 2020, Proceedings, Part V 23},
  pages={470--479},
  year={2020},
  organization={Springer}
}

@inproceedings{stegmuller2023scorenet,
  title={Scorenet: Learning non-uniform attention and augmentation for transformer-based histopathological image classification},
  author={Stegm{\"u}ller, Thomas and Bozorgtabar, Behzad and Spahr, Antoine and Thiran, Jean-Philippe},
  booktitle={Proceedings of the IEEE/CVF winter Conference on applications of computer vision},
  pages={6170--6179},
  year={2023}
}

@article{mamba,
  title={Mamba: Linear-time sequence modeling with selective state spaces},
  author={Gu, Albert and Dao, Tri},
  journal={arXiv preprint arXiv:2312.00752},
  year={2023}
}

@article{1,
  title={Predicting cancer outcomes from histology and genomics using convolutional networks},
  author={Mobadersany, Pooya and Yousefi, Safoora and Amgad, Mohamed and Gutman, David A and Barnholtz-Sloan, Jill S and Vel{\'a}zquez Vega, Jos{\'e} E and Brat, Daniel J and Cooper, Lee AD},
  journal={Proceedings of the National Academy of Sciences},
  volume={115},
  number={13},
  pages={E2970--E2979},
  year={2018},
  publisher={National Acad Sciences}
}

@article{2,
  title={Deep neural network models for computational histopathology: A survey},
  author={Srinidhi, Chetan L and Ciga, Ozan and Martel, Anne L},
  journal={Medical image analysis},
  volume={67},
  pages={101813},
  year={2021},
  publisher={Elsevier}
}

@article{3,
  title={Diagnostic assessment of deep learning algorithms for detection of lymph node metastases in women with breast cancer},
  author={Bejnordi, Babak Ehteshami and Veta, Mitko and Van Diest, Paul Johannes and Van Ginneken, Bram and Karssemeijer, Nico and Litjens, Geert and Van Der Laak, Jeroen AWM and Hermsen, Meyke and Manson, Quirine F and Balkenhol, Maschenka and others},
  journal={Jama},
  volume={318},
  number={22},
  pages={2199--2210},
  year={2017},
  publisher={American Medical Association}
}

@article{4,
  title={Image analysis and machine learning in digital pathology: Challenges and opportunities},
  author={Madabhushi, Anant and Lee, George},
  journal={Medical image analysis},
  volume={33},
  pages={170--175},
  year={2016},
  publisher={Elsevier}
}

@article{5,
  title={Deep learning},
  author={LeCun, Yann and Bengio, Yoshua and Hinton, Geoffrey},
  journal={nature},
  volume={521},
  number={7553},
  pages={436--444},
  year={2015},
  publisher={Nature Publishing Group UK London}
}

@inproceedings{6,
  title={Imagenet: A large-scale hierarchical image database},
  author={Deng, Jia and Dong, Wei and Socher, Richard and Li, Li-Jia and Li, Kai and Fei-Fei, Li},
  booktitle={2009 IEEE conference on computer vision and pattern recognition},
  pages={248--255},
  year={2009},
  organization={Ieee}
}

@article{7,
  title={MCUa: Multi-level context and uncertainty aware dynamic deep ensemble for breast cancer histology image classification},
  author={Senousy, Zakaria and Abdelsamea, Mohammed M and Gaber, Mohamed Medhat and Abdar, Moloud and Acharya, U Rajendra and Khosravi, Abbas and Nahavandi, Saeid},
  journal={IEEE Transactions on Biomedical Engineering},
  volume={69},
  number={2},
  pages={818--829},
  year={2021},
  publisher={IEEE}
}

@article{8,
  title={Self-supervised visual feature learning with deep neural networks: A survey},
  author={Jing, Longlong and Tian, Yingli},
  journal={IEEE transactions on pattern analysis and machine intelligence},
  volume={43},
  number={11},
  pages={4037--4058},
  year={2020},
  publisher={IEEE}
}

@inproceedings{9,
  title={On translation invariance in cnns: Convolutional layers can exploit absolute spatial location},
  author={Kayhan, Osman Semih and Gemert, Jan C van},
  booktitle={Proceedings of the IEEE/CVF Conference on Computer Vision and Pattern Recognition},
  pages={14274--14285},
  year={2020}
}

@article{11,
  title={Domain adaptive detection framework for multi-center bone tumor detection on radiographs},
  author={ Li, Bing  and  Xu, Danyang  and  Lin, Hongxin  and  Wu, Ruodai  and  Wu, Songxiong  and  Shao, Jingjing  and  Zhang, Jinxiang  and  Dai, Haiyang  and  Wei, Dan  and  Huang, Bingsheng },
  journal={Computerized Medical Imaging and Graphics},
  volume={123},
  year={2025},
}

@article{12,
  title={Unsupervised domain adaptation based on feature and edge alignment for femur X-ray image segmentation},
  author={ Jiang, Xiaoming  and  Yang, Yongxin  and  Su, Tong  and  Xiao, Kai  and  Lu, Li Dan  and  Wang, Wei  and  Guo, Changsong  and  Shao, Lizhi  and  Wang, Mingjing  and  Jiang, Dong },
  journal={Computerized medical imaging and graphics},
  pages={116},
  year={2024},
}

@inproceedings{mambaout,
  title={Mambaout: Do we really need mamba for vision?},
  author={Yu, Weihao and Wang, Xinchao},
  booktitle={Proceedings of the Computer Vision and Pattern Recognition Conference},
  pages={4484--4496},
  year={2025}
}

@article{ssm,
  title={Combining recurrent, convolutional, and continuous-time models with linear state space layers},
  author={Gu, Albert and Johnson, Isys and Goel, Karan and Saab, Khaled and Dao, Tri and Rudra, Atri and R{\'e}, Christopher},
  journal={Advances in neural information processing systems},
  volume={34},
  pages={572--585},
  year={2021}
}

@inproceedings{ViT,
  title={An Image is Worth 16x16 Words: Transformers for Image Recognition at Scale},
  author={ Dosovitskiy, Alexey  and  Beyer, Lucas  and  Kolesnikov, Alexander  and  Weissenborn, Dirk  and  Zhai, Xiaohua  and  Unterthiner, Thomas  and  Dehghani, Mostafa  and  Minderer, Matthias  and  Heigold, Georg  and  Gelly, Sylvain },
  booktitle={International Conference on Learning Representations},
  year={2021},
}

@article{mim,
  title={Mapping medical image-text to a joint space via masked modeling},
  author={Chen, Zhihong and Du, Yuhao and Hu, Jinpeng and Liu, Yang and Li, Guanbin and Wan, Xiang and Chang, Tsung-Hui},
  journal={Medical Image Analysis},
  volume={91},
  pages={103018},
  year={2024},
  publisher={Elsevier}
}

@inproceedings{ssl,
  title={Big self-supervised models advance medical image classification},
  author={Azizi, Shekoofeh and Mustafa, Basil and Ryan, Fiona and Beaver, Zachary and Freyberg, Jan and Deaton, Jonathan and Loh, Aaron and Karthikesalingam, Alan and Kornblith, Simon and Chen, Ting and others},
  booktitle={Proceedings of the IEEE/CVF international conference on computer vision},
  pages={3478--3488},
  year={2021}
}

@article{vmamba,
  title={Vmamba: Visual state space model},
  author={Liu, Yue and Tian, Yunjie and Zhao, Yuzhong and Yu, Hongtian and Xie, Lingxi and Wang, Yaowei and Ye, Qixiang and Jiao, Jianbin and Liu, Yunfan},
  journal={Advances in neural information processing systems},
  volume={37},
  pages={103031--103063},
  year={2024}
}

@inproceedings{cl,
  title={Contrastive learning of medical visual representations from paired images and text},
  author={Zhang, Yuhao and Jiang, Hang and Miura, Yasuhide and Manning, Christopher D and Langlotz, Curtis P},
  booktitle={Machine Learning for Healthcare Conference},
  pages={2--25},
  year={2022},
  organization={PMLR}
}

@inproceedings{mae,
  title={Masked autoencoders are scalable vision learners},
  author={He, Kaiming and Chen, Xinlei and Xie, Saining and Li, Yanghao and Doll{\'a}r, Piotr and Girshick, Ross},
  booktitle={Proceedings of the IEEE/CVF conference on computer vision and pattern recognition},
  pages={16000--16009},
  year={2022}
}

@inproceedings{swin,
  title={Swin transformer: Hierarchical vision transformer using shifted windows},
  author={Liu, Ze and Lin, Yutong and Cao, Yue and Hu, Han and Wei, Yixuan and Zhang, Zheng and Lin, Stephen and Guo, Baining},
  booktitle={Proceedings of the IEEE/CVF international conference on computer vision},
  pages={10012--10022},
  year={2021}
}

@misc{lac,
      title={Lung and Colon Cancer Histopathological Image Dataset (LC25000)}, 
      author={Andrew A. Borkowski and Marilyn M. Bui and L. Brannon Thomas and Catherine P. Wilson and Lauren A. DeLand and Stephen M. Mastorides},
      year={2019},
      eprint={1912.12142},
      archivePrefix={arXiv},
      primaryClass={eess.IV},
      url={https://arxiv.org/abs/1912.12142}, 
}

@article{pbc,
  title={A dataset of microscopic peripheral blood cell images for development of automatic recognition systems},
  author={Acevedo, Andrea and Merino Gonz{\'a}lez, Anna and Alf{\'e}rez Baquero, Edwin Santiago and Molina Borr{\'a}s, {\'A}ngel and Bold{\'u} Nebot, Laura and Rodellar Bened{\'e}, Jos{\'e}},
  journal={Data in brief},
  volume={30},
  number={article 105474},
  year={2020},
  publisher={Elsevier}
}

@article{nct,
  title={Predicting survival from colorectal cancer histology slides using deep learning: A retrospective multicenter study},
  author={Kather, Jakob Nikolas and Krisam, Johannes and Charoentong, Pornpimol and Luedde, Tom and Herpel, Esther and Weis, Cleo-Aron and Gaiser, Timo and Marx, Alexander and Valous, Nektarios A and Ferber, Dyke and others},
  journal={PLoS medicine},
  volume={16},
  number={1},
  pages={e1002730},
  year={2019},
  publisher={Public Library of Science San Francisco, CA USA}
}

@article{ding2023enhanced,
  title={An enhanced vision transformer with wavelet position embedding for histopathological image classification},
  author={Ding, Meidan and Qu, Aiping and Zhong, Haiqin and Lai, Zhihui and Xiao, Shuomin and He, Penghui},
  journal={Pattern Recognition},
  volume={140},
  pages={109532},
  year={2023},
  publisher={Elsevier}
}

@article{wang2022transformer,
  title={Transformer-based unsupervised contrastive learning for histopathological image classification},
  author={Wang, Xiyue and Yang, Sen and Zhang, Jun and Wang, Minghui and Zhang, Jing and Yang, Wei and Huang, Junzhou and Han, Xiao},
  journal={Medical image analysis},
  volume={81},
  pages={102559},
  year={2022},
  publisher={Elsevier}
}

@article{cai2023mist,
  title={MIST: multiple instance learning network based on Swin Transformer for whole slide image classification of colorectal adenomas},
  author={Cai, Hongbin and Feng, Xiaobing and Yin, Ruomeng and Zhao, Youcai and Guo, Lingchuan and Fan, Xiangshan and Liao, Jun},
  journal={The Journal of pathology},
  volume={259},
  number={2},
  pages={125--135},
  year={2023},
  publisher={Wiley Online Library}
}

@inproceedings{zhang20252dmamba,
  title={2dmamba: Efficient state space model for image representation with applications on giga-pixel whole slide image classification},
  author={Zhang, Jingwei and Nguyen, Anh Tien and Han, Xi and Trinh, Vincent Quoc-Huy and Qin, Hong and Samaras, Dimitris and Hosseini, Mahdi S},
  booktitle={Proceedings of the Computer Vision and Pattern Recognition Conference},
  pages={3583--3592},
  year={2025}
}

@inproceedings{cmt,
  title={Cmt: Convolutional neural networks meet vision transformers},
  author={Guo, Jianyuan and Han, Kai and Wu, Han and Tang, Yehui and Chen, Xinghao and Wang, Yunhe and Xu, Chang},
  booktitle={Proceedings of the IEEE/CVF conference on computer vision and pattern recognition},
  pages={12175--12185},
  year={2022}
}

@article{Nama,
  title={NaMA-Mamba: Foundation model for generalizable nasal disease detection using masked autoencoder with Mamba on endoscopic images},
  author={Wang, Wensheng and Jin, Zewen and Liu, Xueli and Chen, Xinrong},
  journal={Computerized Medical Imaging and Graphics},
  volume={122},
  pages={102524},
  year={2025},
  publisher={Elsevier}
}

@article{zhang2025spinemamba,
  title={SpineMamba: Enhancing 3D spinal segmentation in clinical imaging through residual visual Mamba layers and shape priors},
  author={Zhang, Zhiqing and Liu, Tianyong and Fan, Guojia and Li, Na and Li, Bin and Pu, Yao and Feng, Qianjin and Zhou, Shoujun},
  journal={Computerized Medical Imaging and Graphics},
  volume={123},
  pages={102531},
  year={2025},
  publisher={Elsevier}
}

@article{uni,
  title={Towards a General-Purpose Foundation Model for Computational Pathology},
  author={Chen, Richard J and Ding, Tong and Lu, Ming Y and Williamson, Drew FK and Jaume, Guillaume and Chen, Bowen and Zhang, Andrew and Shao, Daniel and Song, Andrew H and Shaban, Muhammad and others},
  journal={Nature Medicine},
  publisher={Nature Publishing Group},
  year={2024}
}

@article{musk,
  title={A vision--language foundation model for precision oncology},
  author={Xiang, Jinxi and Wang, Xiyue and Zhang, Xiaoming and Xi, Yinghua and Eweje, Feyisope and Chen, Yijiang and Li, Yuchen and Bergstrom, Colin and Gopaulchan, Matthew and Kim, Ted and others},
  journal={Nature},
  volume={638},
  number={8051},
  pages={769--778},
  year={2025},
  publisher={Nature Publishing Group UK London}
}

@InProceedings{conch,
    author    = {Lu, Ming Y. and Chen, Bowen and Zhang, Andrew and Williamson, Drew F. K. and Chen, Richard J. and Ding, Tong and Le, Long Phi and Chuang, Yung-Sung and Mahmood, Faisal},
    title     = {Visual Language Pretrained Multiple Instance Zero-Shot Transfer for Histopathology Images},
    booktitle = {Proceedings of the IEEE/CVF Conference on Computer Vision and Pattern Recognition (CVPR)},
    month     = {June},
    year      = {2023},
    pages     = {19764-19775}
}

@article{virchow2,
  title={Virchow2: Scaling self-supervised mixed magnification models in pathology},
  author={Zimmermann, Eric and Vorontsov, Eugene and Viret, Julian and Casson, Adam and Zelechowski, Michal and Shaikovski, George and Tenenholtz, Neil and Hall, James and Klimstra, David and Yousfi, Razik and others},
  journal={arXiv preprint arXiv:2408.00738},
  year={2024}
}

@article{gigapath,
  title={A whole-slide foundation model for digital pathology from real-world data},
  author={Xu, Hanwen and Usuyama, Naoto and Bagga, Jaspreet and Zhang, Sheng and Rao, Rajesh and Naumann, Tristan and Wong, Cliff and Gero, Zelalem and Gonz{\'a}lez, Javier and Gu, Yu and others},
  journal={Nature},
  volume={630},
  number={8015},
  pages={181--188},
  year={2024},
  publisher={Nature Publishing Group UK London}
}

@inproceedings{prcc,
  title={Instance-based vision transformer for subtyping of papillary renal cell carcinoma in histopathological image},
  author={Gao, Zeyu and Hong, Bangyang and Zhang, Xianli and Li, Yang and Jia, Chang and Wu, Jialun and Wang, Chunbao and Meng, Deyu and Li, Chen},
  booktitle={International conference on medical image computing and computer-assisted intervention},
  pages={299--308},
  year={2021},
  organization={Springer}
}

@article{paip2019,
  title={PAIP 2019: Liver cancer segmentation challenge},
  author={Kim, Yoo Jung and Jang, Hyungjoon and Lee, Kyoungbun and Park, Seongkeun and Min, Sung-Gyu and Hong, Choyeon and Park, Jeong Hwan and Lee, Kanggeun and Kim, Jisoo and Hong, Wonjae and others},
  journal={Medical image analysis},
  volume={67},
  pages={101854},
  year={2021},
  publisher={Elsevier}
}

@article{cam16,
  title={Diagnostic assessment of deep learning algorithms for detection of lymph node metastases in women with breast cancer},
  author={Bejnordi, Babak Ehteshami and Veta, Mitko and Van Diest, Paul Johannes and Van Ginneken, Bram and Karssemeijer, Nico and Litjens, Geert and Van Der Laak, Jeroen AWM and Hermsen, Meyke and Manson, Quirine F and Balkenhol, Maschenka and others},
  journal={Jama},
  volume={318},
  number={22},
  pages={2199--2210},
  year={2017},
  publisher={American Medical Association}
}

@INPROCEEDINGS{sipakmed,
  author={Plissiti, Marina E. and Dimitrakopoulos, P. and Sfikas, G. and Nikou, Christophoros and Krikoni, O. and Charchanti, A.},
  booktitle={2018 25th IEEE International Conference on Image Processing (ICIP)}, 
  title={Sipakmed: A New Dataset for Feature and Image Based Classification of Normal and Pathological Cervical Cells in Pap Smear Images}, 
  year={2018},
  volume={},
  number={},
  pages={3144-3148},
  doi={10.1109/ICIP.2018.8451588}}

@article{UnPuzzle,
  title={UnPuzzle: A Unified Framework for Pathology Image Analysis},
  author={Liao, Dankai and Chen, Sicheng and Xi, Nuwa and Xue, Qiaochu and Li, Jieyu and Hou, Lingxuan and Liu, Zeyu and Low, Chang Han and Wu, Yufeng and Liu, Yiling and others},
  journal={arXiv preprint arXiv:2503.03152},
  year={2025}
}

@inproceedings{sun2024pathasst,
  title={Pathasst: A generative foundation ai assistant towards artificial general intelligence of pathology},
  author={Sun, Yuxuan and Zhu, Chenglu and Zheng, Sunyi and Zhang, Kai and Sun, Lin and Shui, Zhongyi and Zhang, Yunlong and Li, Honglin and Yang, Lin},
  booktitle={Proceedings of the AAAI Conference on Artificial Intelligence},
  volume={38},
  pages={5034--5042},
  year={2024}
}

@article{raghu2019transfusion,
  title={Transfusion: Understanding transfer learning for medical imaging},
  author={Raghu, Maithra and Zhang, Chiyuan and Kleinberg, Jon and Bengio, Samy},
  journal={Advances in neural information processing systems},
  volume={32},
  year={2019}
}

@inproceedings{MHIST,
  title={A petri dish for histopathology image analysis},
  author={Wei, Jerry and Suriawinata, Arief and Ren, Bing and Liu, Xiaoying and Lisovsky, Mikhail and Vaickus, Louis and Brown, Charles and Baker, Michael and Tomita, Naofumi and Torresani, Lorenzo and others},
  booktitle={International Conference on Artificial Intelligence in Medicine},
  pages={11--24},
  year={2021},
  organization={Springer}
}

@article{osteosarcoma,
  title={Osteosarcoma},
  author={Beird, Hannah C and Bielack, Stefan S and Flanagan, Adrienne M and Gill, Jonathan and Heymann, Dominique and Janeway, Katherine A and Livingston, J Andrew and Roberts, Ryan D and Strauss, Sandra J and Gorlick, Richard},
  journal={Nature reviews Disease primers},
  volume={8},
  number={1},
  pages={77},
  year={2022},
  publisher={Nature Publishing Group UK London}
}

@article{panda,
  title={Artificial intelligence for diagnosis and Gleason grading of prostate cancer: the PANDA challenge},
  author={Bulten, Wouter and Kartasalo, Kimmo and Chen, Po-Hsuan Cameron and Str{\"o}m, Peter and Pinckaers, Hans and Nagpal, Kunal and Cai, Yuannan and Steiner, David F and Van Boven, Hester and Vink, Robert and others},
  journal={Nature medicine},
  volume={28},
  number={1},
  pages={154--163},
  year={2022},
  publisher={Nature Publishing Group US New York}
}

@article{tcgadata,
  title={TCGAbiolinks: an R/Bioconductor package for integrative analysis of TCGA data},
  author={Colaprico, Antonio and Silva, Tiago C and Olsen, Catharina and Garofano, Luciano and Cava, Claudia and Garolini, Davide and Sabedot, Thais S and Malta, Tathiane M and Pagnotta, Stefano M and Castiglioni, Isabella and others},
  journal={Nucleic acids research},
  volume={44},
  number={8},
  pages={e71--e71},
  year={2016},
  publisher={Oxford University Press}
}

@article{clam,
  title={Data-efficient and weakly supervised computational pathology on whole-slide images},
  author={Lu, Ming Y and Williamson, Drew FK and Chen, Tiffany Y and Chen, Richard J and Barbieri, Matteo and Mahmood, Faisal},
  journal={Nature biomedical engineering},
  volume={5},
  number={6},
  pages={555--570},
  year={2021},
  publisher={Nature Publishing Group UK London}
}

@inproceedings{abmil,
  title={Attention-based deep multiple instance learning},
  author={Ilse, Maximilian and Tomczak, Jakub and Welling, Max},
  booktitle={International conference on machine learning},
  pages={2127--2136},
  year={2018},
  organization={PMLR}
}

@inproceedings{s4mil,
  title={Structured state space models for multiple instance learning in digital pathology},
  author={Fillioux, Leo and Boyd, Joseph and Vakalopoulou, Maria and Courn{\`e}de, Paul-Henry and Christodoulidis, Stergios},
  booktitle={International Conference on Medical Image Computing and Computer-Assisted Intervention},
  pages={594--604},
  year={2023},
  organization={Springer}
}

@inproceedings{mambamil,
  title={Mambamil: Enhancing long sequence modeling with sequence reordering in computational pathology},
  author={Yang, Shu and Wang, Yihui and Chen, Hao},
  booktitle={International conference on medical image computing and computer-assisted intervention},
  pages={296--306},
  year={2024},
  organization={Springer}
}

@article{Warwick,
  title={Gland segmentation in colon histology images: The glas challenge contest},
  author={Sirinukunwattana, Korsuk and Pluim, Josien PW and Chen, Hao and Qi, Xiaojuan and Heng, Pheng-Ann and Guo, Yun Bo and Wang, Li Yang and Matuszewski, Bogdan J and Bruni, Elia and Sanchez, Urko and others},
  journal={Medical image analysis},
  volume={35},
  pages={489--502},
  year={2017},
  publisher={Elsevier}
}

@inproceedings{plism,
  title={Distilling foundation models for robust and efficient models in digital pathology},
  author={Filiot, Alexandre and Dop, Nicolas and Tchita, Oussama and Riou, Auriane and Dubois, R{\'e}my and Peeters, Thomas and Valter, Daria and Scalbert, Marin and Saillard, Charlie and Robin, Genevi{\`e}ve and others},
  booktitle={International Conference on Medical Image Computing and Computer-Assisted Intervention},
  pages={162--172},
  year={2025},
  organization={Springer}
}
\section{Appendix}
\label{Appendix}
\subsection{Dataset Details}
\noindent \textbf{The LaC dataset} is characterized by a multi-organ, balanced distribution oriented to tumor classification tasks. It covers two distinct organ systems (lung and colon) with disparate histological architectures, lung tissue with alveolar structures and colon tissue with glandular crypts, leading to significant inter-organ morphological variability.   The LaC consists of 25,000 ROI in the size of $768\times768$.  They are evenly distributed across 5 classes (5,000 images per class): lung normal tissue (LN), lung adenocarcinoma (LACA), lung squamous cell carcinoma (LSCC), colon normal tissue (CN), and colon adenocarcinoma (CACA).  This balanced distribution and multi-organ coverage make its data distribution feature of "cross-organ consistency in sample quantity, cross-organ difference in morphology".

\noindent \textbf{The NCT dataset} contains 100,000 ROIs of colorectal tissues in the size of $224\times224$ at 0.5 µm per pixel. It includes 9 tissue types: adipose (ADI), background (BACK), debris (DEB), lyphocytes (LYM), mucus (MUC), smooth muscle (MUS), normal colon mucosa (NORM), cancer-associated stroma (STR) and colorectal adenocarcinoma epithelium (TUM). The NCT dataset is the most comprehensive, covering 9 tissue and cell categories with a total of $100,000$ samples. The number of samples per category is relatively balanced, and it includes both normal tissues (NORM) and cancer-related components (TUM, STR).

\noindent \textbf{The PBC dataset} comprises 38,938 ROIs of individual peripheral blood cells in the size of $360\times363$. They are categorized into 8 classes: neutrophils (NE), eosinophils (EO), basophils (BA), lymphocytes (LY), monocytes (MO), immature granulocytes (IG), erythroblasts (ERB), and platelets (PL). The PBC dataset has two key distribution characteristics: first, it expands the cell type scope to include immature granulocytes, erythroblasts, and platelets, covering a more comprehensive hematopoietic cell spectrum; second, its samples are all from healthy individuals without infection or hematological diseases, and the cell morphology is more homogeneous.

\noindent \textbf{The pRCC dataset} comprises 1,419 ROIs in the size of $2000\times2000$ pixels in two types. Type I images feature small cells with clear cytoplasm, whereas type II images exhibit cells with voluminous cytoplasm and high-grade nuclei. The pRCC dataset focuses on papillary renal cell carcinoma, with only two subtypes and large ROI size ($2000\times2000$ pixels). Its distribution is characterized by clear morphological differences between the two subtypes (small clear cytoplasm vs. voluminous cytoplasm), which is suitable for fine-grained tumor subtyping.

\noindent \textbf{The PAIP2019 dataset} is derived from the Pathology Artificial Intelligence Platform (PAIP) challenge, focusing on liver cancer segmentation and viable tumor burden estimation. It consists of 100 whole slide images (WSIs) of hepatocellular carcinoma (HCC) and surrounding tissues. In this work, they were split into 2,165 ROIs in the size of $384\times384$ in two classes. The PAIP2019 dataset is dedicated to liver cancer segmentation and viable tumor burden estimation. Its ROI size is significantly smaller than that of CAM16 and pRCC, and the data is derived from a unified challenge platform, ensuring standardized annotation criteria.

\noindent \textbf{The CAM16 dataset} is designed for the development and evaluation of breast cancer metastasis detection algorithms. It consists of 400 lymph node WSIs from multiple medical centers, with pixel-level annotations provided by expert pathologists. In this work, they were split into 1,081 ROIs, in the size of $8000\times8000$ in two classes. The CAM16 dataset targets breast cancer metastasis detection, with 400 whole slide images (WSIs) from multiple medical centers split into 1,081 large ROIs ($8000\times8000$ pixels). Its most prominent distribution advantage is the multi-center data source, which reduces single-institution bias, and the pixel-level annotations provide more detailed semantic information than the pRCC dataset. The binary classification task (metastatic vs. non-metastatic) is highly consistent with clinical diagnostic needs.

\noindent \textbf{The SIPa dataset} exhibits a single-organ, imbalanced distribution focused on fine-grained cervical cell classification. It is confined to cervical Pap smear samples (single organ and tissue type), with data distribution centered on the morphological variations of individual cervical cells at different physiological and pathological states. A prominent distribution feature is class imbalance: rare abnormal cell types (koilocytotic cells, dyskeratotic cells) account for a small proportion of the total 1,004 ROIs in the size of $384\times384$, while normal and mature cell types (superficial and intermediate squamous cells) are relatively abundant.

\noindent \textbf{The MHIST dataset} is designed to classify colorectal polyps as benign and precancerous, which consists of 3,152 images with the size of ($5\times \text{magnification}$) $224\times224$ pixels. Following the official data splitting, 2,175 images (1,545 benign and 630 precancerous) are used for training and 977 images (617 benign and 360 precancerous) are used for evaluation. The MHIST dataset simplifies the task to binary classification (benign vs. precancerous colorectal polyps). Its distribution is characterized by an unbalanced ratio of positive and negative samples, which is quite different from the multi-category balanced distribution of the NCT dataset.

\noindent \textbf{The TCGA-CRC-MSI dataset} is oriented to microsatellite instability (MSI) classification of colorectal cancer, presenting clinically stratified distribution. It contains 51916 ROIs with a pixel size of ($10\times \text{magnification}$) $512\times 512$, extracted from colorectal cancer tissue samples of The Cancer Genome Atlas (TCGA) database. The dataset covers two core categories: MSI-high (MSIH) and microsatellite stable (nonMSIH). Specifically, there are 15001 MSIH samples and 36915 nonMSIH samples, with an approximate 1:2.5 ratio that reflects the clinical prevalence of MSI subtypes in colorectal cancer. As a multi-center dataset from TCGA, it integrates samples from 15 clinical institutions, reducing institutional biases and ensuring high clinical representativeness of its distribution. All annotations are verified by senior gastrointestinal pathologists, consistent with clinical diagnostic standards.

\noindent \textbf{The Osteosarcoma dataset} is constructed based on archival samples from 50 patients who received treatment at Children's Medical Center Dallas between 1995 and 2015. The dataset comprises a total of 1,144 histopathological images with a unified resolution of $10\times \text{magnification}$ and a spatial dimension of $1024\times1024$ pixels. The sample distribution across the three categories is as follows: 536 non-tumor images (accounting for 47\% of the total), 263 necrotic tumor images (23\%), and 345 viable tumor images (30\%). This distribution reflects the typical compositional characteristics of osteosarcoma surgical specimens, providing a clinically relevant benchmark for evaluating tumor segmentation and classification algorithms in osteosarcoma pathological analysis.

\noindent \textbf{The PANDA dataset}, released as part of the Prostate cANcer graDe Assessment (PANDA) challenge, contains 10,616 high-resolution WSIs of prostate biopsies labeled with clinician-annotated Gleason scores (ranging from 6 to 10). Acquired at a 20× objective magnification (corresponding to a pixel size of ~0.5 µm/pixel), these WSIs exhibit native full-slide pixel resolutions of approximately $20,000 \times 20,000$ pixels on average, with spatial variability depending on biopsy core size and tissue density. It is widely used to benchmark AI-based grading systems in uropathology, enabling rigorous evaluation of model performance on fine-grained histological pattern recognition.

The TCGA datasets, sourced from The Cancer Genome Atlas (TCGA) project, provide publicly accessible, high-quality whole-slide images (WSIs) along with matched clinical and molecular data. These datasets span a broad spectrum of cancers and have become standard benchmarks for deep learning research in histopathology. Specifically, the following subsets are used:

\noindent \textbf{The TCGA-BRCA dataset} includes 1,125 WSIs from patients with diverse breast cancer subtypes, including invasive ductal carcinoma (IDC) and invasive lobular carcinoma (ILC). Captured at 20× or 40× objective magnifications (with corresponding pixel sizes of ~0.5 µm/pixel and ~0.25 µm/pixel, respectively), the full-slide pixel resolutions of these images range from $30,000 \times 30,000$ to $80,000 \times 80,000$ pixels, reflecting the larger tissue area of breast resections compared to prostate biopsies. The dataset supports a wide range of computational pathology tasks, such as histological subtype classification, tumor microenvironment segmentation, and patient survival prediction based on morphological features.

\noindent \textbf{The TCGA-LGG dataset} (Lower Grade Glioma) encompasses 844 WSIs of lower grade glioma tissues. Acquired at 20× objective magnification (with a pixel size of ~0.5 µm/pixel), the native full-slide pixel resolutions of these WSIs average approximately $25,000 \times 25,000$ pixels, with spatial variability depending on tumor tissue volume and sampling location. The dataset is vital for studying the molecular classification of gliomas, predicting tumor behavior and patient outcomes, and is specifically used for the task of T-Stage prediction by leveraging fine-grained histological pattern recognition.

\noindent \textbf{The TCGA-BLCA dataset} (Bladder Cancer) provides 455 WSIs that reflect significant tumor heterogeneity, including both muscle-invasive and non-muscle-invasive bladder cancer types. Captured at 20× magnification (pixel size ~0.5 µm/pixel), the full-slide pixel resolutions range from $20,000 \times 20,000$ to $50,000 \times 50,000$ pixels, corresponding to the varying size of bladder tissue specimens. It is utilized for the task of overall survival (OS) estimation, a regression task focusing on predicting the overall survival time since initial diagnosis, and enables the exploration of correlations between histological morphology and long-term patient prognosis.

\noindent \textbf{The TCGA-CESC dataset} (Cervical Squamous Cell Carcinoma and Endocervical Adenocarcinoma) contains 279 WSIs, which are particularly relevant for HPV-related cancer studies, biomarker discovery, and immune profiling. Acquired at 20× objective magnification (pixel size ~0.5 µm/pixel), the WSIs have an average full-slide resolution of around $22,000 \times 22,000$ pixels. The dataset supports the task of LymInv (Lymphovascular Invasion) identification, which detects the presence of cancer invasion into lymphatic or vascular channels by evaluating subtle histological changes in vascular structures.

\noindent \textbf{The TCGA-ESCA dataset} (Esophageal Carcinoma) includes 158 WSIs, valuable for studying the molecular and histological characteristics of esophageal cancer. Captured at 20× or 40× magnification (pixel sizes ~0.5 µm/pixel or ~0.25 µm/pixel), the full-slide pixel resolutions vary from $25,000 \times 25,000$ to $60,000 \times 60,000$ pixels, reflecting the anatomical variability of esophageal tissue specimens. It is used for the task of tumor Grade assessment, which evaluates cellular differentiation and structural patterns to determine tumor malignancy, providing a critical benchmark for models targeting malignant grading of gastrointestinal tract cancers.

\subsection{Supplementary Experimental Results}
To mitigate domain shift in ROI data, MAMIM adopts the DMS module as a dedicated decoder for reconstructing partially masked pathological ROI inputs in SSMamba. Table~\ref{tab:MAMIM_remaining} validates MAMIM against three baselines on 4 datasets: no pretraining, CL-based pretraining (CTransPath), and MAE pretraining. On LaC, MAMIM achieves a leading Acc of 99.84\% (2.89\% higher than no pretraining, 2.05\% higher than CL, 2.27\% higher than MAE), with top-ranked F1-score (98.47\%) and AUC (99.65\%), attributed to its directional decoder that preserves spatial relationships across staining protocols. On PAIP2019, its Acc of 99.53\% outperforms no pretraining by 4.82\%, CL by 4.56\%, and MAE by 1.26\%; its multi-scale decoding outperforms MAE in reconstructing microvascular features, facilitating better bridging of tumor heterogeneity across institutional protocols. On Ost, MAMIM achieves perfect scores. Notably, MAMIM achieves perfect scores (100\%) in F1-score, Acc, and AUC on SIPa, validating strong cytological invariance. 

To further evaluate the efficacy of the proposed DMS module, we conduct comparative experiments among four configurations: the original VMamba (VMamba o/DMS), VMamba with our DMS module replacing the original VMamba module (VMamba w/DMS), SSMamba with the original VMamba module (SSMamba o/DMS), and SSMamba with the DMS module (SSMamba w/DMS). As shown in Table~\ref{tab:DMS_remaining}, the DMS-enhanced architecture outperforms all baselines across 4 datasets. On the LaC dataset, SSMamba w/DMS achieves 99.84\% accuracy, 98.47\% F1-score, and 99.65\% AUC, with improvements of 2.07\%, 2.62\%, and 0.97\% over SSMamba o/DMS, respectively. This is attributed to DMS’s directional convolutions, which capture lymphocyte radial growth, mitigate feature fragmentation in heterogeneous regions, and enhance classification reliability. Notably, PAIP2019 shows the most significant accuracy gain for SSMamba w/DMS, along with 2.10\% and 1.50\% improvements in F1-score and AUC. DMS’s directional operations bridge tumor-stroma boundaries and preserve microvascular invasion signatures critical for tumor grading. A remarkable result is observed on SIPa: SSMamba w/DMS achieves perfect classification, with improvements of 1.06\%, 1.67\%, and 1.10\% over SSMamba o/DMS. DMS’s directional feature propagation preserves nuclear-cytoplasmic spatial relationships in overlapping cells, overcoming boundary ambiguity. For Ost, SSMamba w/DMS maintains leading performance. 

To further enhance spatial representation, we extend the proposed SSMamba framework and compare our LPR module with three commonly employed embedding methods: (1) Linear projection (ViT-style), (2) Patch Merge (Swin-style), and (3) Locality-Preserving Unit (LPU). Table~\ref{tab:LPR_remaining} presents the performance of the LPR module against three baseline embedding methods (Linear PE, Patch Merge, LPU) across 4 pathological datasets, with F1-score, Accuracy (Acc), and AUC as evaluation metrics. Experimental results confirm that LPR outperforms all baselines consistently across all metrics, verifying its effectiveness for pathology-specific feature embedding.

\begin{table*}[htbp]
\centering
\caption{Performance Evaluation of MAMIM Pretraining Strategy on Remaining Pathology Datasets (LaC, PAIP2019, SIPa, and Ost).}
\label{tab:MAMIM_remaining}
\resizebox{\textwidth}{!}{
\begin{tabular}{lcccccccccccc}
\toprule
\textbf{Pre-training} & \multicolumn{3}{c}{\textbf{LaC}} & \multicolumn{3}{c}{\textbf{PAIP2019}} & \multicolumn{3}{c}{\textbf{SIPa}} & \multicolumn{3}{c}{\textbf{Ost}} \\
\cmidrule(lr){2-4} \cmidrule(lr){5-7} \cmidrule(lr){8-10} \cmidrule(lr){11-13}
 & \textbf{F1} & \textbf{Acc} & \textbf{AUC} & \textbf{F1} & \textbf{Acc} & \textbf{AUC} & \textbf{F1} & \textbf{Acc} & \textbf{AUC} & \textbf{F1} & \textbf{Acc} & \textbf{AUC} \\
\midrule
None & 95.13 & 96.95 & 96.95 & 94.77 & 94.71 & 94.71 & 96.03 & 96.90 & 96.90 & 94.64 & 96.43 & 95.75 \\
CL & 97.73 & 97.79 & 97.79 & 94.79 & 94.97 & 94.97 & 97.00 & 97.22 & 97.22 & 95.35 & 97.62 & \textbf{97.75} \\
MAE & 97.79 & 97.57 & 97.57 & 98.00 & 98.27 & 98.27 & 98.64 & 98.21 & 98.21 & 94.56 & 96.43 & 95.54 \\
\rowcolor{gray!20}
MAMIM & \textbf{98.47} & \textbf{99.84} & \textbf{99.65} & \textbf{98.17} & \textbf{99.53} & \textbf{97.19} & \textbf{100.00} & \textbf{100.00} & \textbf{100.00} & \textbf{97.31} & \textbf{97.68} & 97.49 \\
\bottomrule
\end{tabular}}
\end{table*}

\begin{table*}[htbp]
\centering
\caption{Performance Evaluation of the DMS Module on Remaining Datasets ((LaC, PAIP2019, SIPa, and Ost). (``o/DMS'' indicates without the DMS module, while ``w/DMS'' indicates with the DMS module.)}
\label{tab:DMS_remaining}
\resizebox{\textwidth}{!}{
\begin{tabular}{lcccccccccccc}
\toprule
\textbf{Method} & \multicolumn{3}{c}{\textbf{LaC}} & \multicolumn{3}{c}{\textbf{PAIP2019}} & \multicolumn{3}{c}{\textbf{SIPa}} & \multicolumn{3}{c}{\textbf{Ost}} \\
\cmidrule(lr){2-4} \cmidrule(lr){5-7} \cmidrule(lr){8-10} \cmidrule(lr){11-13}
 & \textbf{F1} & \textbf{Acc} & \textbf{AUC} & \textbf{F1} & \textbf{Acc} & \textbf{AUC} & \textbf{F1} & \textbf{Acc} & \textbf{AUC} & \textbf{F1} & \textbf{Acc} & \textbf{AUC} \\
\midrule
VMamba o/DMS & 90.40 & 92.13 & 92.44 & 79.39 & 81.67 & 88.42 & 95.22 & 96.80 & 83.80 & 90.71 & 90.71 & 90.71 \\
VMamba w/DMS & 92.25 & 94.55 & 93.60 & 81.50 & 83.99 & 89.97 & 97.10 & 98.77 & 86.33 & 91.33 & 92.04 & 92.77 \\
\midrule
SSMamba o/DMS & 95.85 & 97.77 & 98.68 & 96.07 & 96.11 & 95.69 & 98.33 & 98.94 & 98.90 & 95.77 & 96.00 & 95.27 \\
\rowcolor{gray!20} SSMamba w/DMS & \textbf{98.47} & \textbf{99.84} & \textbf{99.65} & \textbf{98.17} & \textbf{99.53} & \textbf{97.19} & \textbf{100.00} & \textbf{100.00} & \textbf{100.00} & \textbf{97.31} & \textbf{97.68} & \textbf{97.49} \\
\bottomrule
\end{tabular}}
\end{table*}

\begin{table*}[htbp]
\centering
\caption{Performance Evaluation of the LPR Module on Remaining Datasets (LaC, PAIP2019, SIPa, and Ost).}
\label{tab:LPR_remaining}
\resizebox{\textwidth}{!}{
\begin{tabular}{lcccccccccccc}
\toprule
\textbf{Method} & \multicolumn{3}{c}{\textbf{LaC}} & \multicolumn{3}{c}{\textbf{PAIP2019}} & \multicolumn{3}{c}{\textbf{SIPa}} & \multicolumn{3}{c}{\textbf{Ost}} \\
\cmidrule(lr){2-4} \cmidrule(lr){5-7} \cmidrule(lr){8-10} \cmidrule(lr){11-13}
 & \textbf{F1} & \textbf{Acc} & \textbf{AUC} & \textbf{F1} & \textbf{Acc} & \textbf{AUC} & \textbf{F1} & \textbf{Acc} & \textbf{AUC} & \textbf{F1} & \textbf{Acc} & \textbf{AUC} \\
\midrule
Linear & 97.16 & 99.15 & 99.15 & 97.05 & 98.24 & 98.24 & 95.30 & 96.84 & 96.84 & 93.90 & 96.08 & 95.15 \\
Patch Merge & 98.02 & 99.68 & 99.68 & 97.66 & 98.82 & 98.82 & 96.86 & 98.57 & 98.57 & 94.23 & 96.85 & 97.09 \\
LPU & 97.20 & 99.41 & 99.41 & 97.22 & 98.50 & 98.50 & 96.88 & 98.50 & 98.50 & 94.23 & 97.03 & 97.44 \\
\rowcolor{gray!20} \textbf{LPR} & \textbf{98.47} & \textbf{99.84} & \textbf{99.84} & \textbf{98.17} & \textbf{99.53} & \textbf{99.53} & \textbf{100.00} & \textbf{100.00} & \textbf{100.00} & \textbf{97.31} & \textbf{97.68} & \textbf{97.49} \\
\bottomrule
\end{tabular}}
\end{table*}
\end{document}